%% file: main.tex
\documentclass{article}
\usepackage{amssymb}

\usepackage[numbers]{natbib}
\usepackage{multicol}
\usepackage{multirow}
\usepackage{amssymb}
\usepackage[bookmarks=true]{hyperref}
\usepackage{xspace}
\usepackage[noend]{algorithmic}
\usepackage{algorithm}
\usepackage{mathtools}
\usepackage{booktabs}
\usepackage[shortlabels]{enumitem}
\usepackage{xcolor}
\usepackage{colortbl}
\usepackage{booktabs}
\usepackage{subcaption}

\input{preamble}

\definecolor{lightgray}{rgb}{0.95,0.95,0.95}
\definecolor{ankitColor}{rgb}{0.5,0,0.5} 

\usepackage[preprint]{corl_2025}

\title{ManiFlow: A General Robot Manipulation Policy \\via Consistency Flow Training}

\author{
Ge Yan$^1$\quad Jiyue Zhu$^{2*}$\quad Yuquan Deng$^{1*}$ \vspace{0.05in} \\
\textbf{Shiqi Yang}$^{2}$\quad 
\textbf{Ri-Zhao Qiu}$^{2}$ \quad
\textbf{Xuxin Cheng}$^2$ \quad
\textbf{Marius Memmel}$^1$ \vspace{0.05in}\\
\textbf{Ranjay Krishna}$^{1,4\dagger}$\quad \textbf{Ankit Goyal}$^{3\dagger}$\quad \textbf{Xiaolong Wang}$^{2\dagger}$\quad
\textbf{Dieter Fox}$^{1,3,4\dagger}$\vspace{0.05in}\\
$^1$University of Washington\enspace
$^2$UC San Diego
\enspace $^3$Nvidia
\enspace $^4$Allen Institute for Artifical Intelligence\vspace{0.05in}\\
$^*$Equal Contribution \enspace $^\dagger$ Equal Advising 
\vspace{0.05in}\\
 \href{https://maniflow-policy.github.io/}{\textsc{\textbf{\color{deepred}maniflow-policy.github.io}}}
}

\begin{document}
\maketitle

\begin{center}
    \vspace{-0.2in}
    \centering
    \captionsetup{type=figure}
    \includegraphics[width=1.0\textwidth]{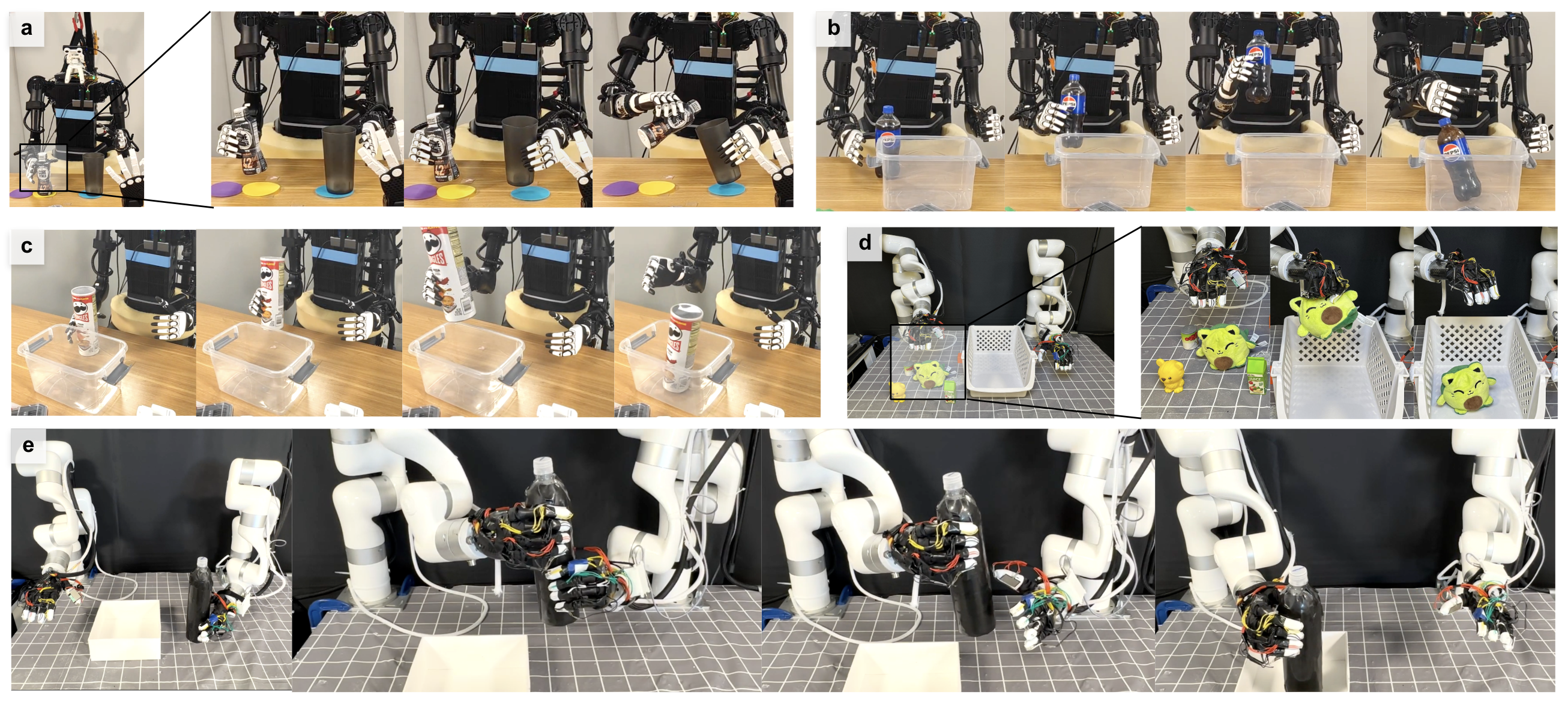}
    \caption{
    We introduce ManiFlow, a flow matching model excelling in complex manipulation tasks, including bimanual dexterous manipulation.
    \textit{a}: Robot autonomously pours water. \textit{b-d}: Robot grasps diverse objects and placing them into containers. \textit{e}: Passing a bottle from one hand to the other.
    }
    \label{fig:teaser}
\end{center}


\begin{abstract}
This paper introduces ManiFlow, a visuomotor imitation learning policy for general robot manipulation that generates precise, high-dimensional actions conditioned on diverse visual, language and proprioceptive inputs. We leverage flow matching with consistency training to enable high-quality dexterous action generation in just 1-2 inference steps. To handle diverse input modalities efficiently, we propose DiT-X, a diffusion transformer architecture with adaptive cross-attention and AdaLN-Zero conditioning that enables fine-grained feature interactions between action tokens and multi-modal observations. ManiFlow demonstrates consistent improvements across diverse simulation benchmarks and nearly doubles success rates on real-world tasks across single-arm, bimanual, and humanoid robot setups with increasing dexterity. The extensive evaluation further demonstrates the strong robustness and generalizability of \ours to novel objects and background changes, and highlights its strong scaling capability with larger-scale datasets. Our website: \ourwebsite.
\end{abstract}


\section{Introduction}

The ability to reliably predict precise and dexterous actions in unstructured environments represents a fundamental challenge in robot learning. Recent advances in diffusion-based policy learning \cite{chi2023diffusion} have significantly enhanced robot capabilities in modeling high-dimensional and multi-modal action distributions. More recently, flow matching \cite{lipman2022flow}, an alternative generative modeling approach, has demonstrated improved performance and training efficiency in policy learning \cite{chisari2024learning, black2410pi0} compared to diffusion-based approaches. 
In spite of these advances, existing flow matching policies \cite{chisari2024learning, black2410pi0,braun2024riemannian, zhang2024affordance} are still limited in efficiency, robustness, and generalizability when performing complex dexterous manipulation tasks in real-world environments. They face challenges in capturing the full complexity of multi-fingered interactions, maintaining temporal coherence across action sequences, generalizing to unseen scenarios, and architectural constraints that insufficiently model multiple data sources inherent in various real-world tasks (e.g., visual, language, proprioception, etc.).

To tackle these challenges, we introduce \ours, a visuomotor imitation model designed to learn robust and generalizable manipulation skills for complex real-world tasks with high dexterity.  
\ours significantly improves previous flow matching policies \cite{zhang2024affordance} through two key contributions. First, we incorporate a consistency training objective into the standard flow matching loss to encourage a more consistent mapping from noisy samples to the target distribution, effectively ``straightening" the flow path. As our experiments show, \ours can generate accurate and dexterous actions in fewer inference steps.  In contrast to previous efforts to reduce inference steps~\cite{prasad2024consistency}, \ours does not rely on any pretrained teacher model, demonstrating better training efficiency. 
Second, we demystify the significance of different time sampling choices with valuable insights and baselines for the flow matching model through comprehensive ablations, indicating the advantage of beta and continuous-time sampling for flow matching and consistency training, as shown in Tab.~\ref{table: ablate time schedular}.

Beyond the consistency flow training process, \ours also improves the model architecture to handle diverse input modalities more effectively with an expressive transformer architecture DiT-X. The DiT-X block builds on the DiT block in image generation \cite{peebles2022scalable} with more effective AdaLN-Zero conditioning for policy learning. Specifically, we use cross-attention layers for high-dimensional visual and language input, with AdaLN-Zero conditioning for low-dimensional inputs like timestep. The learned scale and shift parameters from AdaLN-Zero conditioning are used to adjust the cross-attention input and output features in a selective manner, allowing more efficient and flexible conditioning of multimodal inputs.
Our experiments show that simple yet effective modifications, such as applying AdaLN-Zero conditioning to the cross-attention layers for more adaptive conditioning, significantly improve policy performance compared to previous work, such as MDT \cite{reuss2024multimodal}.

We conduct evaluations across two setups: (1) simulation: 12 tasks in 3 dexterous benchmarks in single-task settings, 48 language-conditioned tasks in multi-task settings, and 4 bimanual dexterous tasks for robustness and generalization test in single-task settings. (2) real-world: 8 challenging tasks across three robot setups with increasing dexterity, including single-arm, bimanual, and humanoid dexterous tasks. We find that \ours consistently improves over diffusion and flow matching policies, both in image-based 2D and pointcloud-based 3D settings.  Specifically, \ours achieves an improvement of 45.6\% and 11.0\% in 12 dexterous tasks with image and pointcloud input, respectively. It further achieves 31.4\% improvement in the multi-task setting. 
Notably, \ours achieves 58\% improvement over the $\pi_0$ model on 4 robustness test tasks and shows superior scaling capability.
Finally, \ours more than doubles the success rate of 3D Diffusion Policy~\cite{ze20243d} across 8 real-world tasks. The key contributions of \ours are three-fold:
\begin{itemize}[leftmargin=*, align=left]
    \item \textbf{High-quality and efficient action generation:} \ours jointly optimizes flow matching with a continuous-time consistency training objective to enforce self-consistency and straightness on learned flow trajectories. This method allows the policy to generate high-dimensional, dexterous actions with high quality using only a few denoising steps, allowing faster inference speed.
    \item \textbf{Efficient multi-modal conditioning:} \ours incorporates DiT-X, a transformer architecture that enhances multi-modal conditioning through adaptive cross-attention layers with learned scale and shift parameters.  This enables selective feature modulation across different input modalities.
    \item \textbf{Real-world robustness and generalizability:} We evaluate \ours on 3 robot setups with increasing dexterity, including challenging bimanual and humanoid dexterous manipulation tasks. \ours consistently shows superior robustness in modeling complex dexterous behavior from limited human demonstrations and significantly improves generalization capability to diverse novel objects and environmental variations.
\end{itemize}

\begin{figure*}[t]
    \centering
    \includegraphics[width=1.0\textwidth]{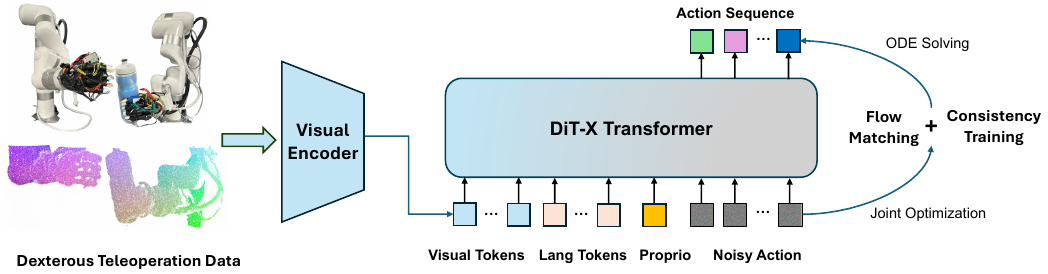}
    \caption{\textbf{Policy Architecture of \ours.} Our system processes 2D or 3D visual observations,
robot state, or language as inputs and outputs a sequence of actions. We leverage a DiT-X transformer architecture to efficiently optimize a flow matching model with a continuous-time consistency training objective, ensuring high-quality action generation for challenging dexterous tasks.}
    \label{fig:main}
\end{figure*}

\section{Method}

\subsection*{Preliminaries: Flow Matching}

We follow \cite{liu2022flow} to define the flow ODE forward process as straight paths between the data distribution and noise. Given a data point $x_1 \sim D$, a noise point $x_0 \sim \mathcal{N}(0,I)$ and timestep $t \sim \mathcal{U}[0,1]$, we define $x_t$ as a linear interpolation between $x_0$ and $x_1$, i.e $x_t = (1-t)\,x_0 + t\,x_1$, and the velocity $v_t$ as the direction from noise to data point: $v_t = x_1 - x_0.$
The flow model $\theta$ is optimized to predict the velocity given a noisy sample $x_t$ at time point $t$. The flow matching loss $\mathcal{L}_\mathrm{FM}(\theta)$ is defined as:
\begin{align}
\mathcal{L}_\mathrm{FM}(\theta) &= \mathbb{E}_{x_0,x_1\sim D}[\|v_\theta(x_t,t) - (x_1 - x_0)\|^2]
\label{eq:flow_training}
\end{align}

\subsection{\ours Training}
\ours goes beyond the basic flow matching model by incorporating a continuous-time consistency training objective and improved time-space sampling strategies, as outlined below.

\subsubsection*{Continuous-time Consistency Training}
\begin{figure*}[t]
    \centering
    \includegraphics[width=1.0\textwidth]{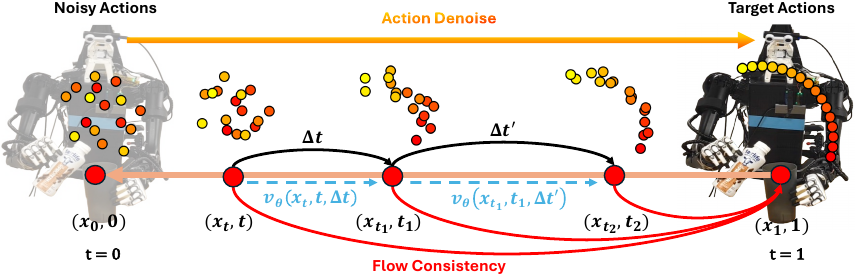}
    \vspace{-0.1in}
    \caption{\textbf{\ours Consistency Training.} Given a flow path that smoothly transforms action to noise, we sample multiple intermediate points via linear interpolation (e.g., $x_{t}$, $x_{t_1}$, and $x_{t_2}$). During training, we learn to map any intermediate point on the flow trajectory back to its origin $x_1$ and ensure the self-consistency of sampled points on the same trajectory.}
    \label{fig:flow_overview}
\end{figure*}

Compared with standard diffusion and flow matching models that require many denoising steps during inference \cite{chi2023diffusion, black2410pi0}, consistency training \cite{song2023consistency} provides an elegant approach to improve generation quality and achieve few-step generation without relying on pre-trained teacher models. The key insight is enforcing the consistency
of partially-noisy data points along an ordinary differential equation (ODE) trajectory to the final target data points. We leverage this principle to jointly optimize the flow matching model with a consistency training objective to enhance the consistency of learned flows and thus generate high-quality action trajectories, as shown in the Fig.~\ref {fig:flow_overview}.

Similar to Shortcut Model~\cite{frans2024one}, we add another argument $\Delta t$ to the flow model $v_t(x_t, t, \Delta t)$, where $\Delta t$ reflects the step size towards the next target point. 
We sample a timestep $t$ from the discretized [0,1] interval and a step size $\Delta t$ from $\mathcal{U}[0,1]$. We define the next timestep $t_1$ as $t+\Delta t$, ensuring that it is bounded in $[0,1]$ via clipping. The velocity $v_{t_1}$ at point $x_{t_1}$ toward a further timestep $t_2$ set as $t_1+\Delta t'$ is predicted as $v_{\theta^-}( x_{t_1}, t_1, \Delta t')$ where $\theta^-$ is the exponential
moving average (EMA) of the flow model.
To enforce consistency between points  $x_t$ and $x_{t_1}$, we first approximate the target data point $\tilde{x}_1 = x_{t_1} + (1- t_1) \cdot v_{t_1}$. We then further estimate the average velocity target $\tilde{v}_{\text{target}}$ from point $x_t$ to $\tilde{x}_1$ as
$\tilde{v}_{\text{target}} = (\tilde{x}_1 - x_t) \mathbin{\boldsymbol{/}} (1-t)$. We enforce consistency by constraining the flow model to predict this estimated velocity target, with the consistency loss $\mathcal{L}_\mathrm{CT}$:
\begin{align}
\mathcal{L}_\mathrm{CT}(\theta) =\mathbb{E}_{t,\Delta t\sim\mathcal{U}[0,1]}\left[\| v_\theta(x_t, t, \Delta t) - \tilde{v}_{\text{target}} \|^2\right]
\label{eq:consistency_training}
\end{align}
We combine flow matching $\mathcal{L}_\mathrm{FM}$ and consistency training losses $\mathcal{L}_\mathrm{CT}$ in \ours training: $\mathcal{L}(\theta) = \mathbb{E}[\|v_\theta(x_t,t, 0) - (x_1 - x_0)\|^2 + \|v_\theta(x_t, t, \Delta t) - \tilde{v}_{\text{target}} \|^2]$, where the third argument ($\Delta t$) in the flow model is set as 0 for the $\mathcal{L}_\mathrm{FM}$ as it estimates local instant velocity~\cite{frans2024one}.
Note that, unlike consistency training \cite{song2023consistency} that operates in discrete time step size $\Delta t$, we sample $\Delta t$ from a continuous distribution to remove the undesirable bias associated with discrete-time objectives and ensure more flexible generation. The EMA model provides essential stabilization \cite{song2023consistency}, with more details in the appendix.

\subsubsection*{Time Space Sampling Strategy}

Time scheduling in generative models significantly impacts learning dynamics and final performance. We evaluate five representative timestep $t$ sampling strategies in flow matching as denoted in Eq.~\ref{eq:flow_training} with visualization and pseudo-code in Fig.~\ref{fig:time_distribution} and Alg.~\ref{alg:time_sampling}: (1) Uniform sampling \cite{liu2022flow}, which draws timesteps uniformly from [0,1] and serves as a straightforward baseline; (2) Logit-normal sampling (lognorm) \cite{atchison1980logistic}, which emphasizes intermediate timesteps through a logit-normal distribution with tunable location and scale parameters; (3) Mode sampling \cite{esser2024scaling}, which allows explicit control over whether to favor midpoint or endpoints during training through a scale parameter $s$; (4) CosMap sampling \cite{nichol2021improved}, which adapts the cosine schedule from diffusion models to the flow matching setting through a specialized mapping function; and (5) Beta distribution sampling \cite{black2024pi_0}, which places more weight on lower timesteps corresponding to noisier actions, with a cutoff threshold $s=0.999$ to avoid sampling timesteps that contribute minimal learning value. As we find in Tab.~\ref{table: ablate time schedular}, while lognorm sampling shows strong performance, the beta distribution's focus on the high-noise regime proves particularly effective for robotic control tasks, outperforming other scheduling strategies across diverse manipulation scenarios. We further ablate the step size choice $\Delta t$ in consistency training, denoted in Eq.~\ref{eq:consistency_training}, and continuous time shows improved performance as shown in Tab.~\ref{table: ablate time schedular}.

\subsection{Perception}

Our 3D visual encoder builds upon \cite{ze20243d} while introducing a key modification to prioritize the preservation of fine-grained geometric information in 3D point cloud representations. 
The key insight is that maintaining detailed spatial relationships throughout the encoding process is crucial for precise manipulation tasks. While previous works like \cite{ze20243d} used max pooling operations to compress point cloud features into a compact representation, we found this compression can lead to loss of important geometric details. Our architecture deliberately avoids such pooling operations, instead preserving point-wise features throughout the network. This design choice allows the encoder to maintain richer spatial relationships and detailed geometric information of the input point cloud, which we found particularly beneficial for tasks requiring precise object interaction and spatial reasoning.

Empirical observations show that scene configuration significantly impacts the optimal point density for representation efficiency. In well-calibrated scenes with cropped points, \ours achieves strong performance with sparse point clouds of 128 points, demonstrating the efficiency of the network. For uncalibrated egocentric views, denser representations of 4096 points are sufficient, suggesting the benefit of increased point density in less structured environments. Note that proper color augmentation is helpful for optimal results without overfitting, as elaborated in the appendix.

\begin{figure}[t]
    \centering
    \vspace{-0.3in}
\includegraphics[width=0.9\textwidth]{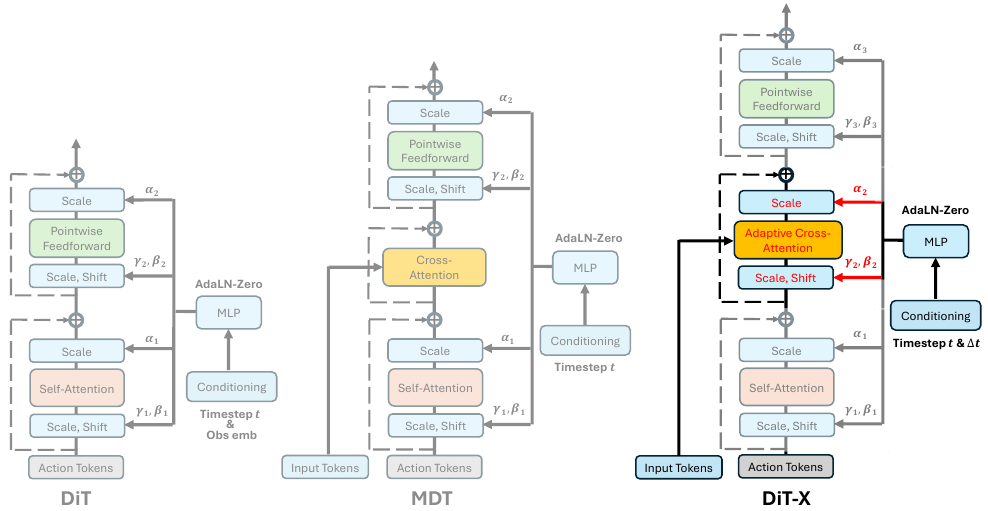}
\caption{\textbf{DiT-X Block.} Unlike DiT (self-attention only) and MDT (basic cross-attention), DiT-X applies AdaLN-Zero conditioning to low-dimensional robot state inputs, and adjusts cross-attention input and output with learned scaling and shift parameters, ensuring adaptive and fine-grained feature interactions between action tokens and multi-modal input tokens. This design enables efficient handling of both low-dimensional control signals and high-dimensional perceptual inputs.}
    \label{fig:dit-x block}
     \vspace{-0.2in}
\end{figure}

\subsection{\ours Policy Architecture}

For the lack of adaptive conditioning in standard cross-attention mechanisms (e.g., MDT \cite{reuss2024multimodal}), we introduce \textbf{DiT-X}, a transformer architecture that effectively processes low-dimensional signals and high-dimensional multi-modal inputs for general robotic control. Our design is motivated by the inherent challenges in generative models for handling diverse input modalities: low-dimensional signals require precise encoding of high-frequency dynamics, visual inputs contain rich spatial-semantic information, and language instructions introduce fine-grained language understanding. We follow the principles below to design an expressive architecture for multi-modality conditioning.

\textbf{Adaptability \& Granularity:} Being capable of generating highly adaptive actions is essential for robot manipulation in a dynamic environment, requiring a reactive adjustment with precision. Additionally, the integration of high-dimensional visual and language features with low-dimensional signal demands fine-grained understanding and adaptive interaction collectively. 
We address this through a dedicated adaptive cross-attention mechanism that enables direct token-level interactions between actions and multi-modal inputs, facilitating precise spatial and semantic alignment.

\begin{wrapfigure}{r}{0.3\linewidth}
\centering
 \vspace{-2em}
\begin{minipage}{\linewidth}
    \captionsetup[subfigure]{justification=centering,font=small}
    \centering
    \includegraphics[width=0.9\linewidth]{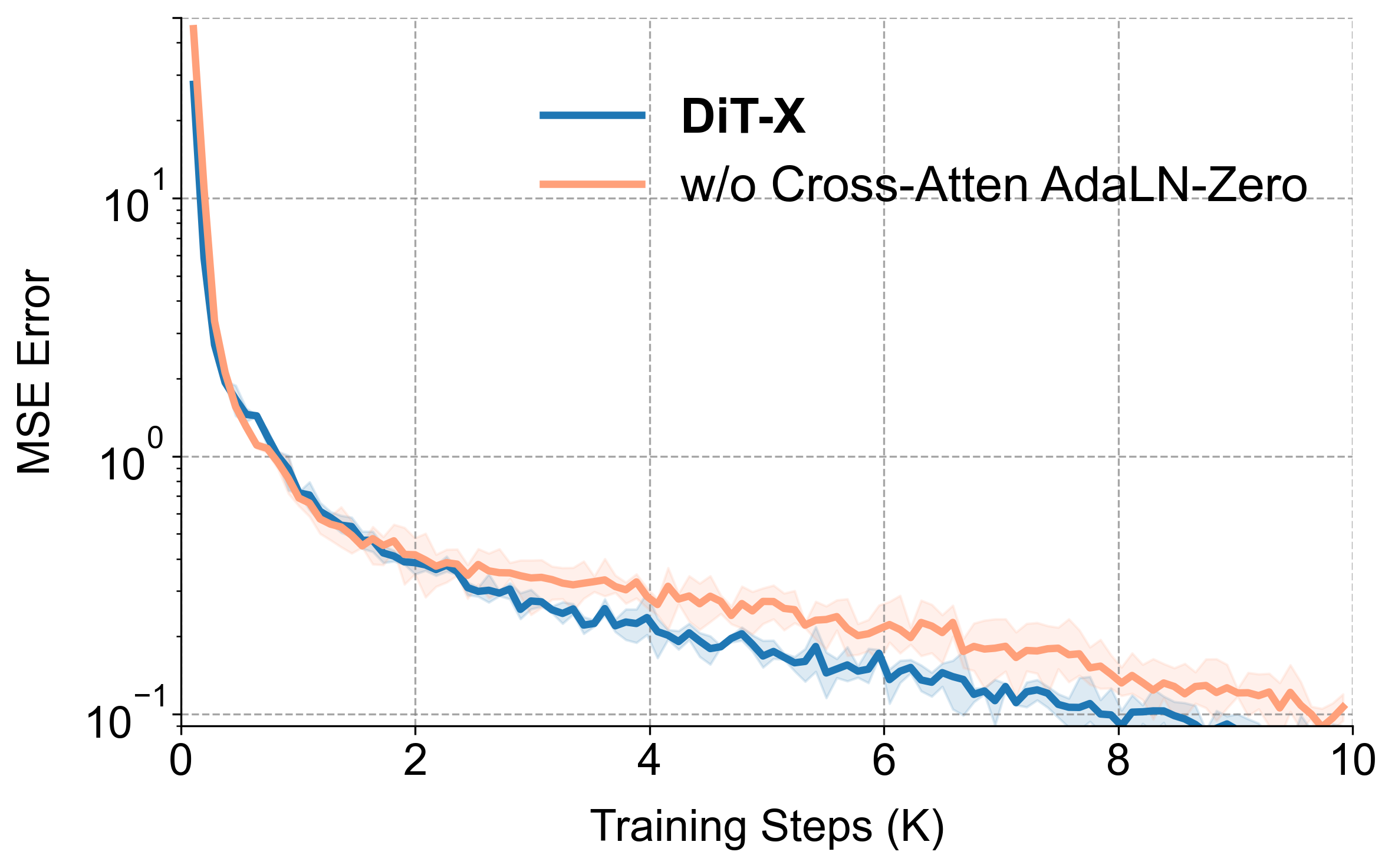}
 \vspace{-1.5ex}
    \label{fig:ditx_component:a}
    \vspace{-1.5ex}
\includegraphics[width=0.9\linewidth]{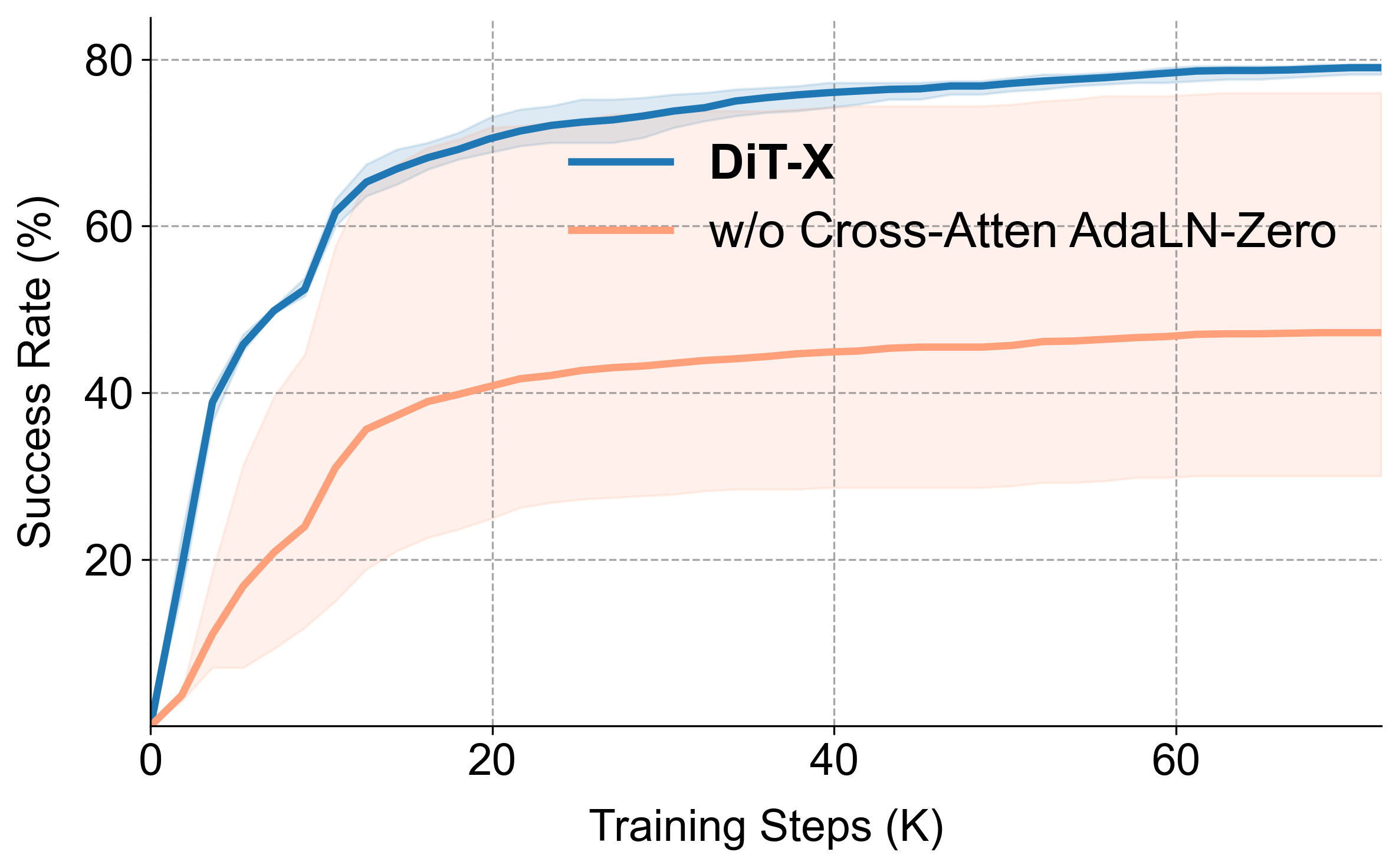}
    \label{fig:ditx_component:b}
\end{minipage}
\captionsetup{font=small}
\caption{Training action error and success rate of DiT-X vs w/o cross-attention AdaLN-zero conditioning in 10 Metaworld tasks with language conditioning.}
\label{fig:ditx_component}
\vspace{-0.6cm}
\end{wrapfigure}

\textbf{DiT-X block with Adaptive Cross-attention Conditioning:} We introduce adaptive cross-attention layers to process visual and language tokens with low-dimensional input. Specifically, given low-dimensional inputs like timesteps, we employ AdaLN-Zero conditioning \cite{peebles2022scalable} to generate conditioning scale and shift parameters ($\alpha, \gamma, \beta$) for dynamic adaptation of network behavior while ensuring stable training through zero initialization. In particular, instead of only applying scale and shift parameters to self-attention and feedforward layers, we also adjust the input and output of cross-attention layers with the same modulation. This design empowers the network to manipulate fine-grained visual and language tokens by scaling them down or up in a selective manner, which is crucial for tasks requiring a precise understanding of visual cues and language instructions. While this introduces a modest computational overhead, the enhanced representational capability proves valuable for complex manipulation tasks.

As shown in Fig.~\ref{fig:ditx_component}, the DiT-X block shows faster convergence during training and better performance than w/o cross-attention AdaLN-Zero conditioning. Furthermore, we provide a detailed illustration of the evolving DiT and MDT architecture baselines in Fig.~\ref{fig:dit-x block}. Our architecture provides greater expressiveness than the DiT and MDT blocks on multi-modality conditioning in Fig.~\ref{fig:DiT_learning_efficiency}.

\input{tables/results_sim_simplified}

\begin{figure*}[t]
    \centering
    \includegraphics[width=0.95
    \textwidth]{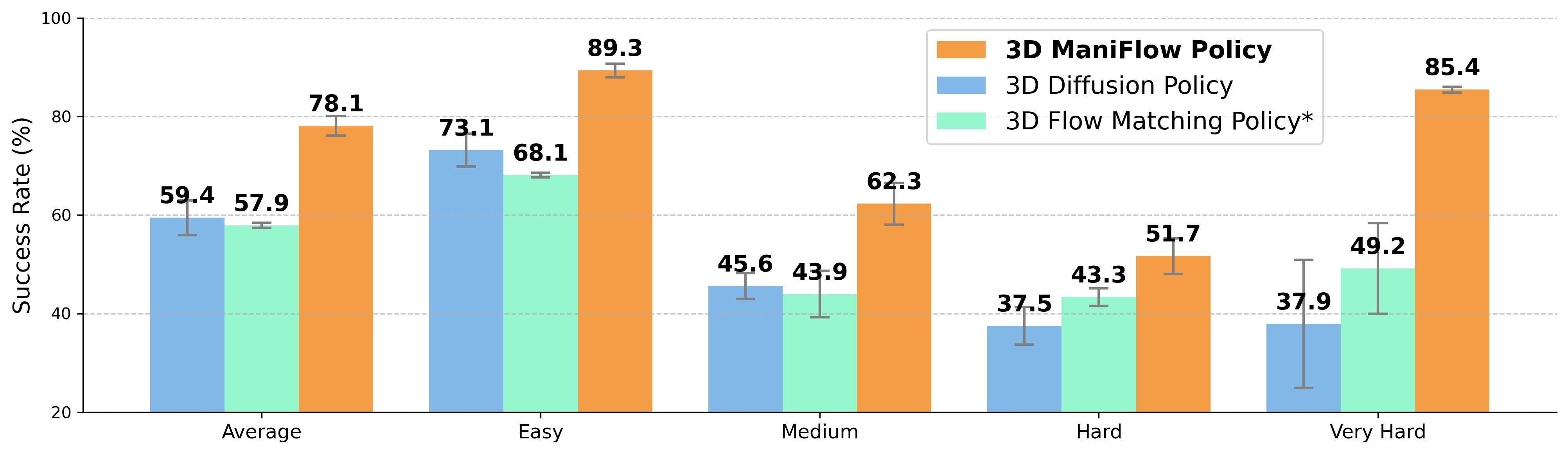}
    \vspace{-0.1in}
    \caption{\textbf{Comparison on language-conditioned multi-task learning on 48 MetaWorld tasks.} \ours achieves superior performance across all difficulty levels
    compared to the 3D diffusion and flow matching policy, with an average \textbf{31.4\%} and \textbf{34.9\%} relative improvement. } 
    \label{fig:metaworld_multitask}
    \vspace{-0.2in}
\end{figure*}

\vspace{-0.1in}
\section{Experiments}
\vspace{-0.1in}

\subsection{Simulation Experiments}
\vspace{-0.05in}

\textbf{Benchmarks:} We select three diverse dexterous manipulation benchmarks (Adroit \cite{Kumar2016thesis}, Dexart \cite{bao2023dexart}, and RoboTwin 1.0 \cite{mu2024robotwin} to comprehensively evaluate \ours in 12 dexterous tasks that assess a wide spectrum of manipulation capabilities.  Furthermore, with the MetaWorld benchmark \cite{yu2020meta} comprising 48 tasks, we specifically focus on the challenging language-conditioned multi-task learning scenario to provide a comprehensive assessment of model performance when conditioning on visual and language input. 
We further use the RoboTwin 2.0 benchmark \cite{chen2025robotwin}
to fully test the policy robustness and generalizability.
More details are provided in the appendix.

\textbf{Baselines:} For 2D image inputs, we compare \ours with diffusion policy~\cite{chi2023diffusion} and flow matching policy~\cite{zhang2024affordance} with the same ResNet-18 encoder \cite{he2016deep}.  For 3D pointcloud-based methods, we primarily compare against 3D Diffusion Policy (DP3)~\cite{ze20243d}, which has demonstrated superior performance over 2D Diffusion Policy across various simulation environments. Since the flow matching policy~\cite{zhang2024affordance} is only image-based in the original paper, we add the same 3D encoder from \cite{ze20243d} to it in order to get a baseline for the 3D-based flow matching model, denoted as 3D Flow Matching Policy*. For the robustness test and scaling experiment on the RoboTwin 2.0 benchmark, we compare with the $\pi_0$ model, which takes multi-view images as input and is fine-tuned on the domain randomized data.

\begin{figure}[t]
    \centering
    \begin{subfigure}{0.5\textwidth}
        \centering
        \includegraphics[width=\textwidth]{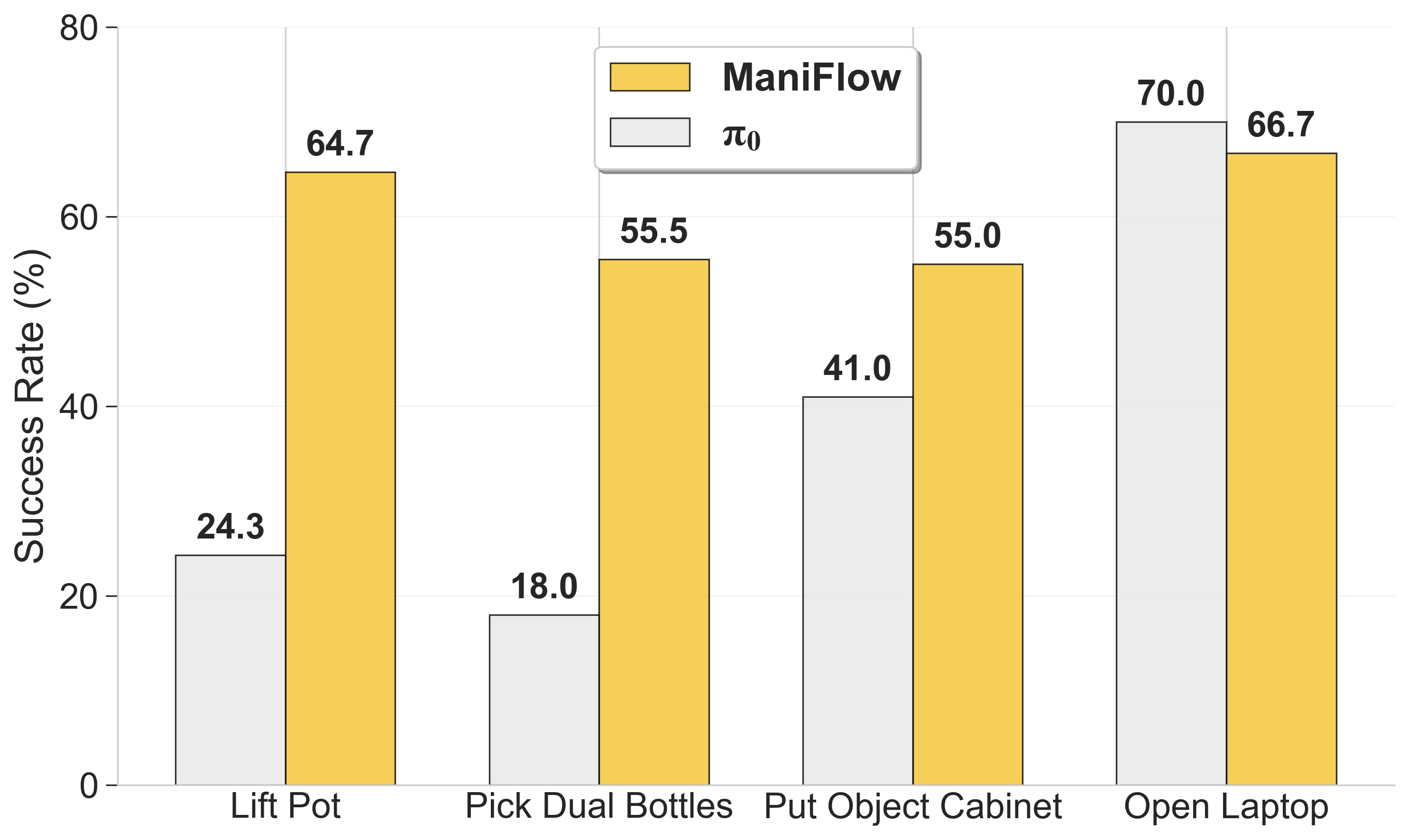}
        \caption{}
        \label{fig:maniflow_efficiency}
    \end{subfigure}
    \hfill
    \begin{subfigure}{0.48\textwidth}
        \centering
        \includegraphics[width=\textwidth]{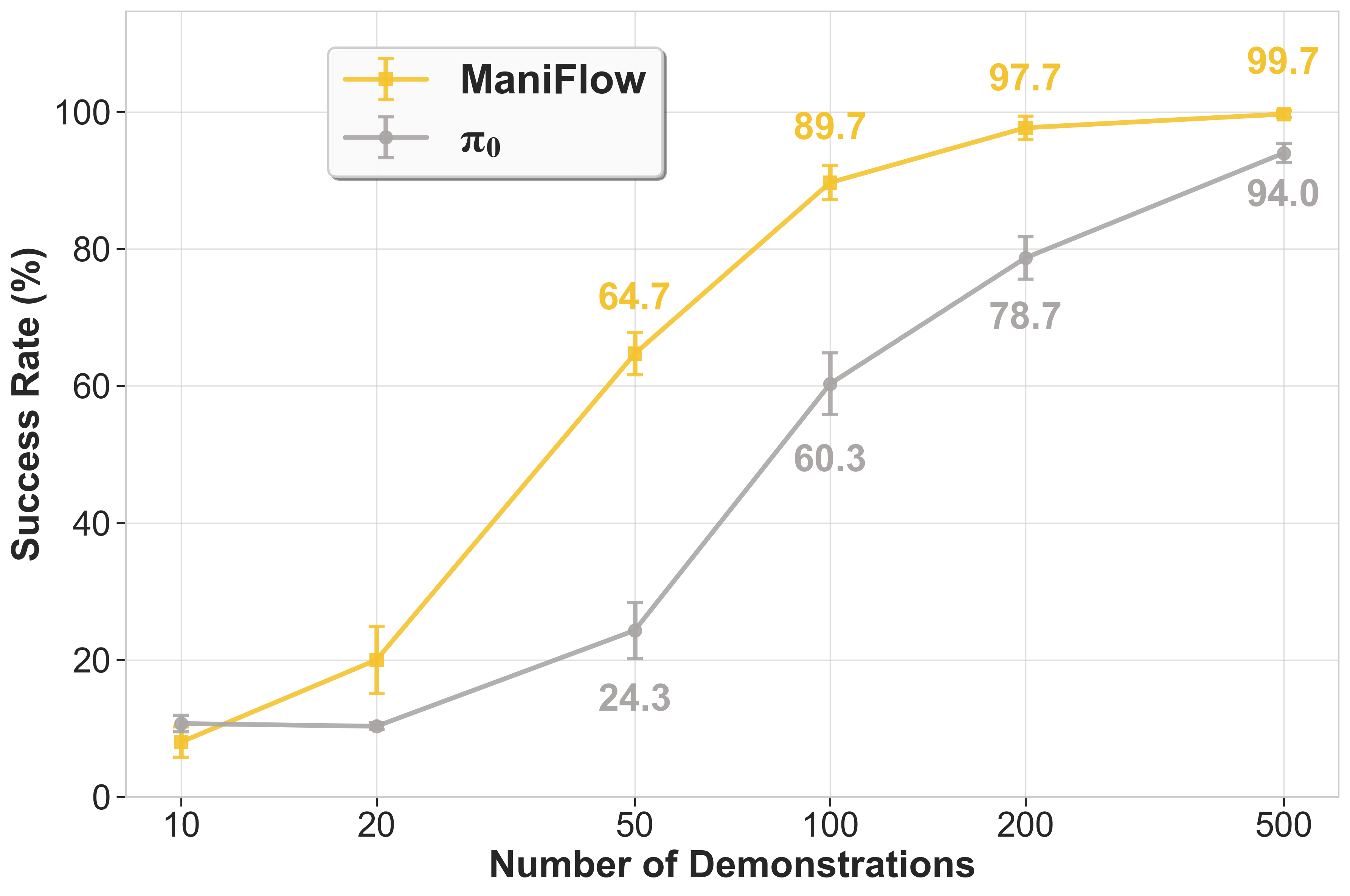}
        \caption{}
        \label{fig:maniflow_scaling}
    \end{subfigure}
    \caption{ \textbf{(a) Efficiency \& Generalization. }We evaluate \ours and $\pi_0$ with 4 bimanual tasks on RoboTwin 2.0 benchmark (Fig.~\ref{fig:robotwin2_tasks}), after training with 50 domain randomized demonstrations per task. Compared to the large-scale pre-trained $\pi_0$ model, \ours shows superior learning efficiency and generalization capability to novel objects and backgrounds, while learning from scratch with pointcloud input.
    \textbf{(b) Scaling Behavior.} Results show the scaling performance on the task "lift pot" with demonstration numbers varying from 10 to 500. \ours consistently outperforms $\pi_0$ on both the low data regime and final scaling to 500 demos, achieving 99.7\% success eventually. 
    }
    \label{fig:maniflow_efficiency_scaling}
\end{figure}

\begin{figure}[t]
    \centering
    \vspace{-0.2cm}
    \includegraphics[width=0.9\textwidth]{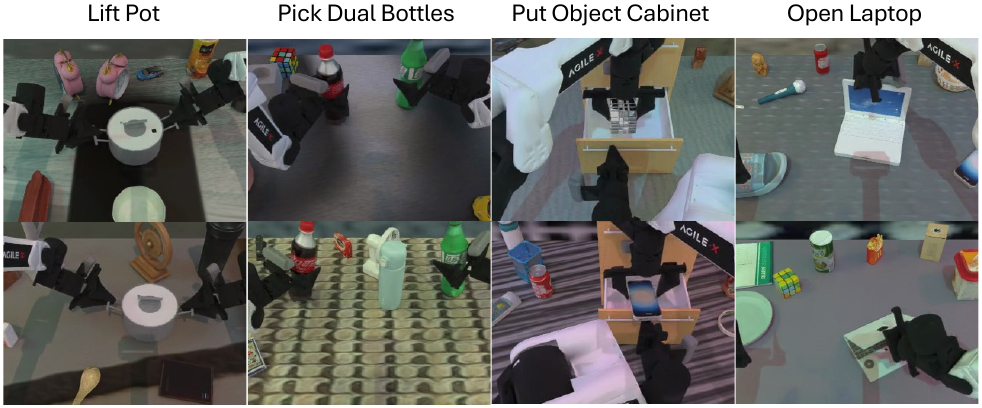}
    \caption{\textbf{Visualization of Domain Randomized Evaluation.} To fully test the robustness and generalizability of our policy, we evaluate both \ours and $\pi_0$ on the RoboTwin 2.0 benchmark with challenging domain randomizations, including 
    cluttered scenes with random distractors, novel objects and diverse background textures, various lighting conditions, and table height changes.
    }
    \label{fig:robotwin2_tasks}
\end{figure}

\begin{figure*}[t!]
    \centering
    \includegraphics[width=1.0\textwidth]{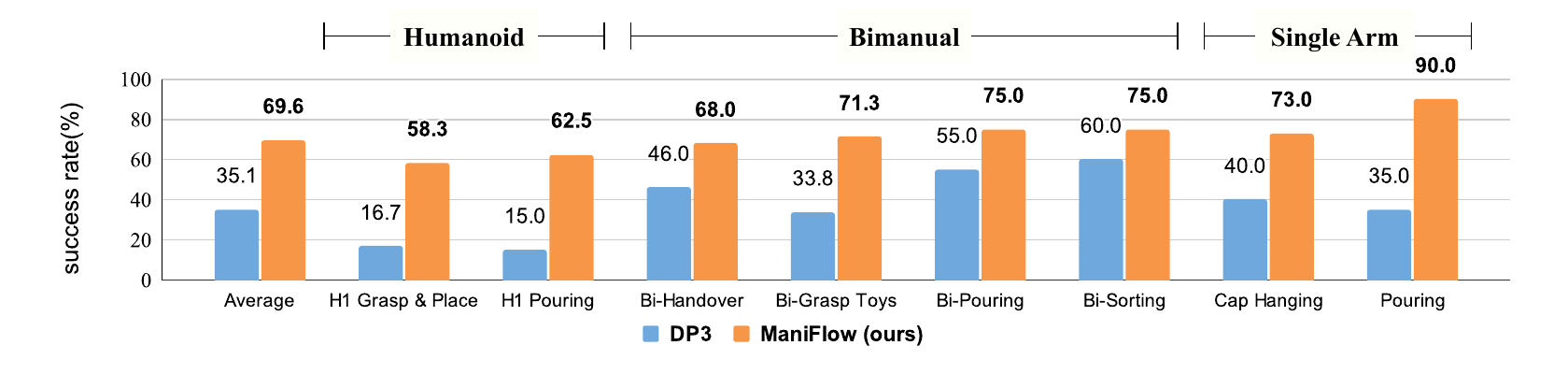}
    
    \vspace{0.1in}
    
    \includegraphics[width=1.0\textwidth]{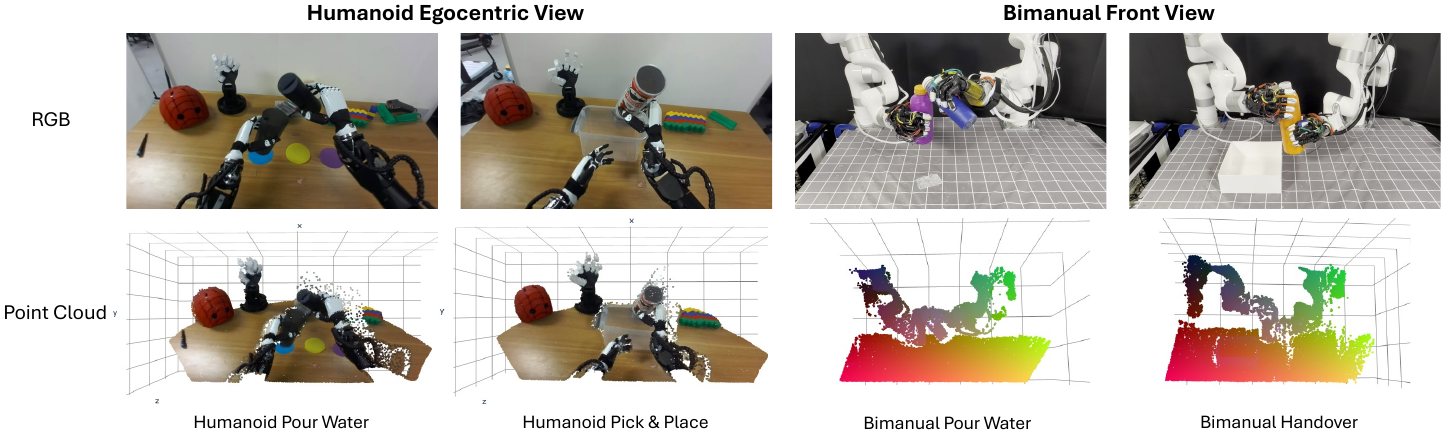}
    
    \caption{\textbf{Real-Robot Results:} (Top) We test 8 real-robot tasks across 3 robot platforms, including Franka with gripper, bimanual xArm with ability hands, and Unitree H1 humanoid with bimanual anthropomorphic hands. \ours succeeds \textbf{69.6\%} on average, almost doubling DP3's performance. \textbf{Visualizations:} (Bottom) 3D point cloud visualizations of sampled 4 real robot tasks.}
    \label{fig:real_world_results}
\end{figure*}

\subsection{Key Findings}

As shown in Tab.~\ref{table:simulation simplified}, \ours outperforms both 2D image and 3D point cloud-based diffusion and flow matching policies on all 3 dexterous benchmarks, with an average 43.4\% and 45.6\% improvement on 2D input, and 15.9\% and 11.0\% improvement on 3D input. \ours further achieves 78.1\% success rate in language-conditioned multi-task learning on 48 MetaWorld tasks, demonstrating 31.4\% and 34.9\% improvement (see Fig.~\ref{fig:metaworld_multitask}). Notably, \ours achieves 58\% improvement over the $\pi_0$ model on 4 bimanual tasks with point cloud input, also demonstrating superior scaling capability. We discuss the key takeaways below and provide further ablations in the appendix.

\textbf{High-quality action generation.} Dexterous manipulation poses a significant challenge in the model's ability to capture high-dimensional behaviors. We observe that \ours consistently achieves higher success rates compared to the 3D diffusion and flow matching policy. This performance advantage is clearly demonstrated in the most challenging bimanual dexterous tasks in the RoboTwin 1.0 benchmark, where \ours achieves a success rate of 61.9\% with only 50 demonstrations, while DP3 achieves 42.7\% success rate (see Tab.~\ref{table:simulation simplified}). The performance gap is particularly notable given the challenging nature of bimanual coordination.

\textbf{Robust visual and language conditioning.} \ours demonstrates better visual conditioning capability than diffusion and flow matching policies for both 2D and 3D visual input. Notably, for the Adroit 3 tasks in Tab.~\ref{table:simulation simplified}, 2D \ours achieves 73.2\% success rate, while both 2D baselines struggle in this benchmark. Additionally, for language conditioning, we evaluate against 3D-based baselines on 48 MetaWorld tasks with multi-task learning in Fig.~\ref{fig:metaworld_multitask}. \ours outperforms 3D diffusion and flow matching baselines on all task difficulty levels
by a large margin: 31.4\% and 34.9\% relative improvement on average, and notable 125\% and 73.6\% on the very hard tasks. 

\textbf{Enhanced performance through DiT-X architecture.} Our experimental results on 10 language-conditioned MetaWorld tasks demonstrate the significant advantages of \ours's DiT-X block over the DiT and MDT architectures.  As shown in Fig.~\ref{fig:DiT_learning_efficiency}, DiT-X achieves faster learning and better final performance on various tasks.  
DiT-X's adaptive cross-attention AdaLN-Zero conditioning mechanism enables more fine-grained interactions between visual features, language instructions, and action sequences, which is crucial for language-conditioned tasks where success depends on a precise understanding of both visual cues and natural language commands. 

\textbf{Learning Efficiency \& Generalization.} As demonstrated in Fig.~\ref{fig:maniflow_efficiency_scaling}(a), \ours achieves superior learning efficiency compared to the fine-tuned $\pi_0$ model across 4 challenging bimanual dexterous tasks on the RoboTwin 2.0 benchmark. Training from scratch with only 50 domain randomized demonstrations per task, \ours substantially outperforms $\pi_0$: 64.7\% vs 24.3\% on \textit{Lift Pot}, 55.5\% vs 18.0\% on \textit{Pick Dual Bottles}, 55.0\% vs 41.0\% on \textit{Put Object Cabinet}, and 66.7\% vs 70.0\% on \textit{Open Laptop}, achieving 58\% relative improvement on average. Beyond learning efficiency, \ours demonstrates robust generalization to environmental variations including novel objects, diverse backgrounds, cluttered scenes with distractors, and varying lighting conditions as shown in Fig.~\ref{fig:robotwin2_tasks}. This combination of efficiency and generalization capability suggests that \ours effectively learns robust and generalizable manipulation skills from limited demonstrations, outperforming even large-scale pre-trained models in challenging unseen scenarios.

\textbf{\ours Scaling Behavior.} \ours exhibits strong scaling capability across different data regimes, as shown on the \textit{lift pot} task in Fig.~\ref{fig:maniflow_efficiency_scaling}(b). Starting from comparable performance at 10 demonstrations ($\sim$10\% for both methods), \ours shows a clear performance advantage in the low-data regime: achieving 64.7\% success rate at 50 demonstrations compared to $\pi_0$'s 24.3\%, and quickly reaching $\sim$90\% success with 100 demonstrations while $\pi_0$ achieves 60.3\%. Notably, \ours demonstrates better data scaling behavior by achieving 97.7\% success with 200 demonstrations, while $\pi_0$ requires 500 demonstrations to reach 94.0\%, still below \ours's 200-demo performance. \ours continues to improve to 99.7\% at 500 demos. The consistent upward scaling trajectory indicates that \ours leverages larger scale demonstration data more effectively than $\pi_0$, suggesting better scaling properties for learning complex dexterous behaviors with more data.

\textbf{\ours excels in few-step inference.} Due to the costly iterative denoising steps, few-step inference is essential for sufficiently fast policy generation in the real world. As shown in Tab.~\ref{table: inference} in the appendix, \ours achieves 63.7\% and 64.5\% success rate using only 1 and 2 inference steps, respectively, compared to 3D Diffusion and Flow Matching Policies using 10 inference steps to achieve 42.7\% and 48.1\% success rate on 5 bimanual dexterous tasks in the RoboTwin benchmark.

\subsection{Real World Experiments}

We evaluate ManiFlow on 8 real-robot tasks across 3 robot setups with increasing dexterity (see Fig.~\ref{fig:real_world_results} and Tab.~\ref{table:real_main_results}). Each setup is evaluated on a unique set of tasks designed to assess ManiFlow’s capabilities across diverse scenarios. We provide an overview of the robot setups in Fig.~\ref{fig:real_world_setup} and task visualizations in the appendix. We compare \ours against DP3, the previous state-of-the-art dexterous manipulation policy. Both \ours and DP3 take point clouds as visual input.  As can be seen, \ours consistently outperforms DP3 by a significant margin: 88.8\%  relative improvement for in-distribution environment configurations and 116.7\% on unseen objects, leading to 98.3\% relative improvement on average. 

\textbf{High Dexterity:} As shown in Tab.~\ref{table:real_main_results}, \ours excels in tasks requiring high dexterity, particularly evident in its performance with anthropomorphic hands on the Unitree H1 humanoid and bimanual setups. \ours demonstrates strong capability in tasks such as pouring, where it must precisely control multi-finger positions to grasp the bottle without missing and aligning the bottle opening with the cup carefully, showing improved success rate from 20\% to 65\% on the humanoid platform. The additional complexity of bimanual coordination, requiring synchronization between two independent dexterous hands, further highlights ManiFlow's superiority. As shown in the handover task that requires the left hand to grasp the bottle first and hand it to the right hand, \ours succeeds on 22 out of 30 runs (73\% success rate) compared to DP3's success on 14 out of 30 runs (47\%). 

\textbf{Generalization:} 
\ours is able to handle unseen object types and geometries (e.g., varying bottle heights, appearances, and shapes) along with changes in the environment without any significant drop in performance (see Tab.~\ref{table:real_main_results}). On the other hand, DP3 often halted mid-motion or failed to recognize and adapt to new objects during task execution. This inability to handle unfamiliar objects was particularly evident when DP3 was tasked with manipulating unseen objects in the Toy Grasping tasks. In contrast, our method was able to adapt to novel objects and successfully executed the tasks with minimal disruption.  Furthermore, \ours demonstrated robustness to changes in the scene, such as distractor objects, cluttered environments, and varying backgrounds. On the other hand, in tasks like Toy Grasping with randomly placed distractors, DP3 showed a tendency to overfit to the specific end-effector trajectories seen during training.

 \begin{figure*}[t!]
    \centering
    \includegraphics[width=1.0\textwidth]{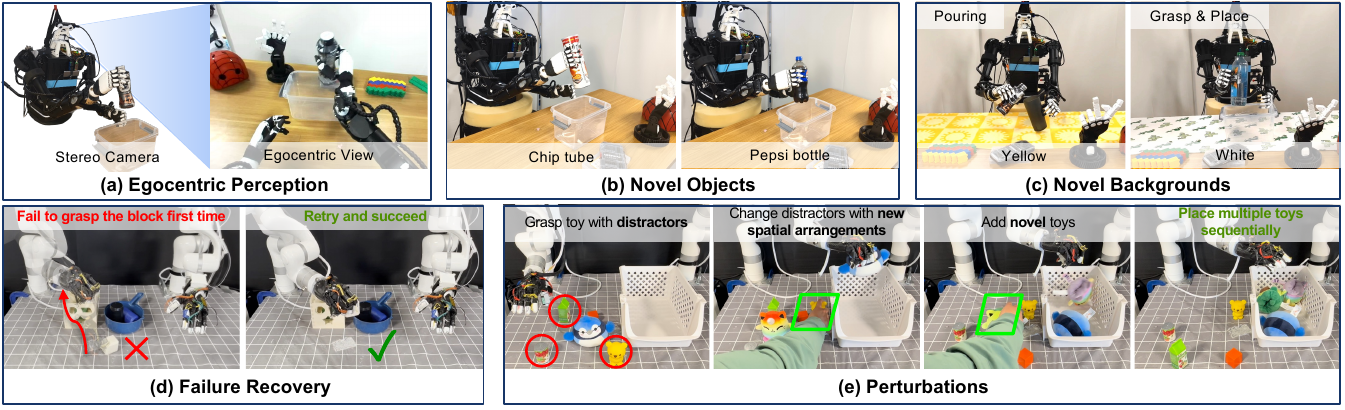}
    \vspace{-0.1in}
    \caption{\textbf{Real World Robustness.} We test the policy robustness with varying perturbations during real-world deployment, such as different egocentric viewpoints, novel objects and backgrounds, recovering from failure, and adding diverse distractors with human perturbed locations. \ours is robust against these perturbations with limited data. Please check our \href{https://maniflow-policy.github.io/}{\textcolor{deepred}{website}} for more details.
    }
    \label{fig:todo}
    \vspace{-0.1in}
\end{figure*}

 \input{tables/results_real}


\section{Related Work}
\vspace{-0.1in}

\textbf{Generative Models for Policy Learning:}
Diffusion models, a family of generative models that iteratively transform random noise into a data sample, have achieved great success in generating
high-resolution images and videos. Owning to this impressive success, they have also been applied in various robotics domains. Notably, Diffusion Policies~\cite{chi2023diffusion} have been effective in modeling multi-modal action distributions. 
Building on them, Consistency Policies~\cite{prasad2024consistency} used a pre-trained diffusion model to distill a student model. By using this two-stage pipeline, they demonstrated faster inference with fewer denoising steps. Recently, flow matching has demonstrated improved performance and training efficiency in policy learning~\cite{zhang2024affordance, chisari2024learning}. However, these methods still face limitations in modeling more complex and high-dimensional dexterous behaviors. We improve upon the flow matching model by using a consistency training objective. \ours shows strong capability in generating high-quality actions with only a few inference steps, demonstrating both robustness and efficiency in challenging dexterous tasks. Notably, \ours can be trained end-to-end in a single run without requiring an additional teacher model, unlike other methods \cite{prasad2024consistency, lu2024manicm, jia2024score} that typically require pre-training models for teacher-student distillation or multiple training stages for inference acceleration, making them computationally expensive and more cumbersome to work with. 

\textbf{Visual Imitation Learning.}
Prior works have shown that visual observations are essential for robots to have an accurate understanding of the environment. While 2D image-based imitation learning policies have been widely adopted due to the simplicity and easy access of RGB images, policies that take in 3D input have demonstrated better performance and generalizability. Recent works~\cite{shridhar2023perceiver, goyal2023rvt, goyal2024rvt, ze2023gnfactor, yan2024dnact, li2025integrating} have shown success in leveraging 3D data for manipulation tasks. However, these methods are typically restricted to low-dimensional 6-DoF end-effector control with coarse temporal keypoints prediction. Hence, they are not suitable for highly dynamic and dexterous tasks. Beyond these methods, 3D Diffuser Actor~\cite{ke20243d} can predict continuous dense actions, but is still restricted to 6-DoF end-effector control and not applicable for high-dimensional dexterous manipulation. 3D Diffusion Policy~\cite{ze20243d} leverages an efficient 3D encoder and achieves superior performance for dexterous tasks. Compared to this line of works, we aim to develop a general robot policy that is capable of learning robust manipulation skills from either 2D or 3D observations.

\textbf{Architecture for Multi-modality Conditioning}
Recent advancements in robotic manipulation have leveraged data from different modalities
to improve robustness and sample efficiency in complex real-world environments. 
Prior works have developed a high-capacity diffusion transformer (DiT) \cite{peebles2023scalable} and applied it to manipulation tasks \cite{dasari2024ingredients}, demonstrating better visual conditioning compared to traditional transformer architecture. 
A related work MDT \cite{reuss2024multimodal} showed improved performance by incorporating cross-attention layers to fuse multimodal conditioning information. \ours builds upon these prior works and improves them further through the DiT-X block.
We add a simple yet effective modification: introducing the AdaLN-Zero conditioning to the cross-attention layer with learned scaling and shift parameters to better manipulate the conditioned network’s features in a selective manner, allowing more flexible and efficient multimodal conditioning.

\section{Conclusion}
\label{sec:conclusion}
In this work, we introduce \ours, a robust and efficient dexterous manipulation model.  \ours improves upon prior flow matching policies by introducing a continuous-time consistency training objective, a superior time sampling strategy, and a novel DiT-X block.  The proposed DiT-X architecture effectively handles diverse input modalities through its dual conditioning mechanisms, enabling strong performance across varied manipulation tasks. Our comprehensive evaluation spanning 64 simulation tasks and 8 real-world scenarios demonstrates \ours's effectiveness, particularly in challenging real-world bimanual dexterous manipulation, where it achieves a 98.3\% relative improvement over existing approaches. 

\section{Limitation}
While \ours demonstrates strong performance across diverse manipulation tasks, there are several promising avenues for future work. The success in real-world robot tasks depends heavily on the quality and diversity of training demonstrations. Incorporating \ours into a reinforcement learning framework could potentially reduce the burden on the demonstration data.  Furthermore, while the design choices for \ours are inspired by dexterous manipulation tasks, none of these are limited to robot manipulation, and we believe that \ours could be equally beneficial for tasks such as navigation or mobile manipulation.  Finally, we only scratched the surface of \ours's multi-modal capabilities, and the incorporation of further modalities such as tactile information or VLM-based conditioning via points, trajectories, or bounding boxes is an interesting extension.


\acknowledgments{
Part of this work was funded by the Army Research Lab and award \#W911NF-24-2-0191.
}



\bibliography{main}  

\newpage
\input{appendix}

\end{document}

%% file: preamble.tex
\usepackage{multirow}
\usepackage{multicol}
\usepackage{graphicx}
\usepackage{xspace}
\usepackage{xcolor}
\usepackage{caption}
\usepackage{wrapfig}
\usepackage{bbding}  
\usepackage{pifont}  
\usepackage{booktabs}
\usepackage{float}
\usepackage{amsmath}  
\usepackage{amssymb}  
\newcommand{\ours}[0]{ManiFlow\xspace}
\newcommand{\ourwebsite}[0]{\href{https://maniflow-policy.github.io/
}{maniflow-policy.github.io
}\xspace}

\usepackage{colortbl}
\definecolor{ourcolor}{HTML}{99e0eb}
\definecolor{ourblue}{HTML}{27a2c3}
\definecolor{tablecolor}{HTML}{eef2f5}
\definecolor{tablecolor2}{HTML}{ffcdb4}
\definecolor{citecolor}{HTML}{fe7b5b}
\definecolor{grey}{rgb}{0.9, 0.9, 0.9}
\usepackage{amssymb}
\usepackage{listings}
\usepackage{adjustbox}
\definecolor{gred}{rgb}{0.859,0.267,0.216}
\definecolor{ggreen}{rgb}{0.059,0.616,0.345}
\definecolor{deepblue}{HTML}{27a2c3}
\definecolor{deepred}{HTML}{fe7b5b}

\newcommand{\dd}[2]{$#1\scriptstyle{\pm#2}$}
\newcommand{\ddbf}[2]{\cellcolor{tablecolor}$\mathbf{#1\scriptstyle{\pm#2}}$}

%% file: tables/results_sim_simplified.tex
\begin{table*}[t]
\centering
\caption{\textbf{Main Simulation Results.} Success rates on 12 dexterous tasks in 3 benchmarks. \ours achieves superior performance on both image and point cloud-based inputs.}
\label{table:simulation simplified}
\vspace{-0.05in}
\resizebox{1.0\textwidth}{!}{%
\begin{tabular}{l|>{\centering\arraybackslash}p{0.6cm}|ccccccc|cc}

\toprule

Algorithm $\backslash$ Task & Obs. & RoboTwin 5 tasks & Adroit 3 tasks & DexArt 4 tasks &   \multicolumn{2}{c}{Average}\\
\midrule
Diffusion Policy & Img & \dd{28.8}{2.3} & \dd{38.1}{2.9} & \dd{53.6}{2.1} & \multicolumn{2}{c}{\dd{39.4}{2.3}}\\
Flow Matching Policy & Img & \dd{27.1}{2.7}  & \dd{39.0}{2.2} & \dd{53.3}{2.4} & \multicolumn{2}{c}{\dd{38.8}{2.5}}\\
\textbf{2D \ours Policy} & Img &\ddbf{46.1}{2.7} & \ddbf{74.3}{1.9} & \ddbf{56.3}{2.3} & \multicolumn{2}{c}{\ddbf{56.5}{2.4}}\\
\midrule  
3D Diffusion Policy & PC & \dd{42.7}{3.3} & \dd{77.8}{2.4} & \dd{60.6}{0.7}  & \multicolumn{2}{c}{\dd{57.4}{2.2}}\\
3D Flow Matching Policy* & PC & \dd{48.1}{6.3} & \dd{77.1}{3.3} & \dd{61.7}{1.1} & \multicolumn{2}{c}{\dd{59.9}{2.8}}\\
\textbf{3D \ours Policy} & PC & \ddbf{61.9}{2.5} &  \ddbf{78.6}{2.3} & \ddbf{63.2}{2.7} &  \multicolumn{2}{c}{\ddbf{66.5}{2.5}}\\
\bottomrule
\end{tabular}}
\vspace{-0.1in}
\end{table*}

%% file: tables/results_real.tex
\begin{table}[H]
\vspace{-2ex}
\centering
\caption{Detailed Comparison of DP3 and ManiFlow on 8 real robot tasks across 3 robot platforms}
\resizebox{0.7\textwidth}{!}{
\begin{tabular}{@{}c|c|cc|cc@{}}
\toprule
\multirow{2}{*}{\textbf{Real Robot Setup}} & \multirow{2}{*}{\textbf{Task}} & \multicolumn{2}{c|}{\textbf{In Distribution}} & \multicolumn{2}{c}{\textbf{Unseen Objects}} \\
\cmidrule(lr){3-4} \cmidrule(lr){5-6}
 & & DP3 & \textbf{ManiFlow} & DP3 & \textbf{ManiFlow} \\
\midrule
\multirow{2}{*}{Humanoid} & Grasp \& Place & 7/40 & \textbf{23/40} & 3/20 & \textbf{12/20} \\
 & Pouring & 4/20 & \textbf{13/20} & 2/20 & \textbf{12/20} \\
\midrule
\multirow{4}{*}{Bimanual} & Handover & 14/30 & \textbf{22/30} & 9/20 & \textbf{12/20} \\
 & Pouring & 21/40 & \textbf{30/40} & 12/20 & \textbf{15/20} \\
 & Toy Grasping & 17/50 & \textbf{37/50} & 7/30 & \textbf{20/30} \\
 & Sorting & 7/10 & \textbf{8/10} & 5/10 & \textbf{7/10} \\
\midrule
\multirow{2}{*}{Single-Arm} & Cap Hanging & 4/10 & \textbf{7/10} & 2/5 & \textbf{4/5} \\
 & Pouring & 5/10 & \textbf{9/10} & 2/10 & \textbf{9/10} \\
\midrule
\multicolumn{2}{c|}{\textbf{Average Success Rate}} & 37.6\% & \textbf{71.0\%} & 31.1\% & \textbf{67.4\%} \\
\bottomrule
\end{tabular}}
\vspace{-3ex}
\label{table:real_main_results}
\end{table}

%% file: appendix.tex
\appendix


\section{Policy Implementation Details}

\begin{center}
    \centering
    \captionsetup{type=figure}
    \includegraphics[width=1.0\textwidth]{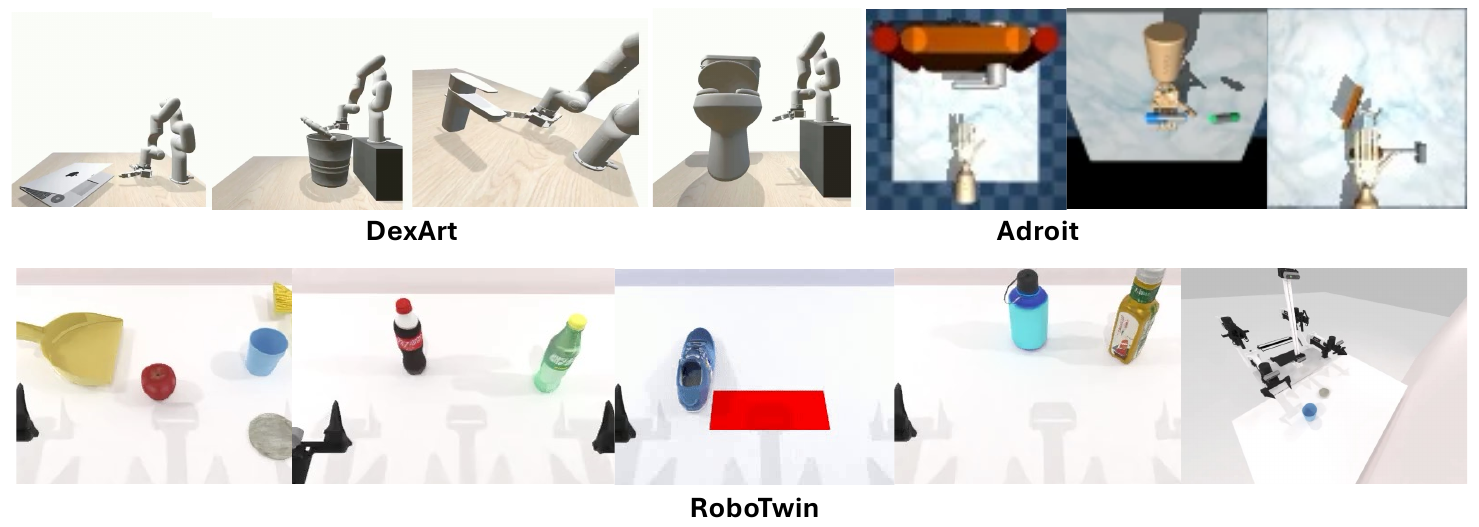}
    \caption{
    Simulation Tasks Visualization. 12 dexterous manipulation tasks, including 4 DexArt tasks, 3 Adroit tasks, and 5 bimanual dexterous RoboTwin 1.0 tasks.
    }
    \label{fig:sim_tasks}
\end{center}

\begin{figure}[htbp]
    \centering
    \includegraphics[width=0.9\textwidth]{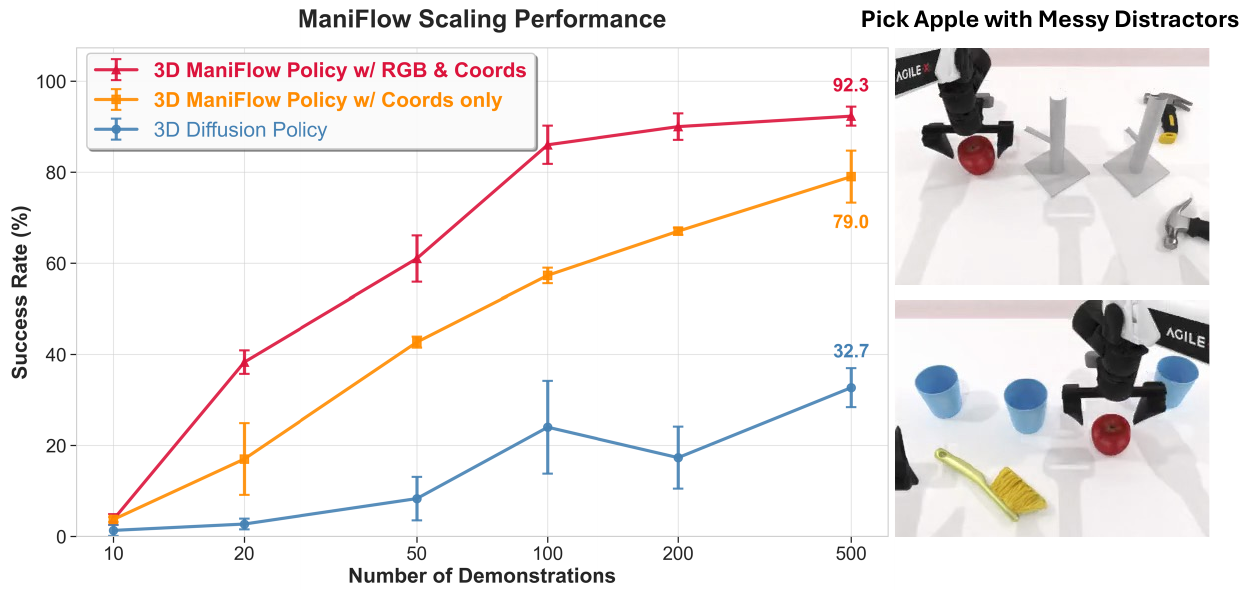}
    \caption{\textbf{Scaling Comparison.} We evaluate 3D ManiFlow Policy and 3D Diffusion Policy across 10 to 500 demos on the \textit{Pick Apple Messy} task from the RoboTwin 1.0 benchmark. ManiFlow achieves a 79.0\% success rate with 500 demonstrations using point cloud coordinates only, significantly outperforming the diffusion baseline at 32.7\%. Adding RGB information further improves performance to 92.3\%, demonstrating superior data efficiency and scaling capability of \ours.}
    \label{fig:maniflow_scaling}
\end{figure}

\subsection{Perception}

ManiFlow is designed as a general robot policy capable of learning robust manipulation skills from either 2D or 3D visual observations. Our experiments demonstrate consistent improvements over baseline methods in both modalities, with 32.9\% relative improvement on 2D image inputs and 16.2\% improvement on 3D point cloud inputs across dexterous manipulation benchmarks. We detail our 2D and 3D visual encoding approaches below.

\textbf{2D Visual Encoding.} For 2D image-based inputs, we train a ResNet-18 encoder from scratch to process RGB images and extract adaptive visual features optimized by policy gradient. The resulting visual tokens are fed into our DiT-X transformer for cross-attention conditioning with action tokens. We apply  a set of image augmentations, including random crop (ratio 0.95), random rotation (±5 degrees), and color jitter (brightness 0.3, contrast 0.4, saturation 0.5, hue 0.08) to improve generalization and robustness. This design choice is primarily for less noisy simulation environments. For in-the-wild real-world environments, we recommend using larger, pre-trained visual encoders to learn more robust and reactive behavior, as demonstrated in UMI \cite{chi2024universal}.

\textbf{3D Visual Encoding.} Our 3D visual encoder builds upon a lightweight pointnet encoder \cite{ze20243d} while removing max pooling operations to preserve point-wise 3D features for fine-grained geometric understanding. We elaborate the key design choices for deploying 3D-based \ours subsequently. 

\textbf{Point Cloud Density.} ManiFlow can learn from varying point cloud densities efficiently. In well-calibrated and cropped scenes, ManiFlow only needs very sparse point clouds with as few as 128 to 256 points. For more cluttered environments, ManiFlow adopts denser 2048 to 4096 points to ensure adequate spatial coverage and preserve important geometric details in complex scenes.

\textbf{Point Cloud Augmentation.} We found that SE3 spatial augmentation is detrimental to performance and do not use it in our training. In most simulation tasks, we use point cloud coordinates only unless specifically noted. However, as demonstrated in Fig.~\ref{fig:maniflow_scaling}, adding color information can substantially improve performance in cluttered environments as it provides rich semantics regarding various objects and surroundings. In real-world experiments, color jitter augmentation becomes essential for generalizing to environment changes and preventing overfitting to specific lighting conditions. We apply the same color jitter parameters as in image augmentation to the RGB in point clouds with 0.2 probability, significantly improving robustness and generalizability in real-world deployment.

\textbf{Learn from Egocentric View.} ManiFlow is applied to both third-person view cameras with static viewpoints and egocentric view with active sensing cameras that have moving viewpoints. For third-person setups, cameras are positioned externally to provide consistent, fixed perspectives of the manipulation workspace, as seen in our real-world bimanual and single-arm experiments. For egocentric setups, such as the humanoid configuration with gimbal-mounted stereo cameras, the visual perspective dynamically changes as the robot's head moves during data collection, requiring the policy to handle varying viewpoints and coordinate head-arm movements simultaneously. 

\textbf{More capable 3D Encoders.} While our current lightweight PointNet-based encoder prioritizes simplicity and efficiency for dexterous manipulation, it may be limited in highly complex in-the-wild scenes that require richer semantic understanding. Future enhancements could address these limitations through two primary directions: (1) integrating pre-trained 3D foundation models \cite{wang2025vggt, zhou2023uni3d} to leverage large-scale geometric and semantic priors for improved generalization to novel objects and environments, and (2) lifting 2D semantic features from vision-language models into 3D space \cite{ze2023gnfactor, yan2024dnact, ke20243d}, to combine our efficient geometric processing with rich semantic understanding. These approaches would strengthen ManiFlow's robustness and adaptability to more challenging real-world scenarios with diverse objects, cluttered environments, and varying lighting conditions.

\subsection{\ours \& Baseline Model Details.}
\textbf{Language Encoding.} For language-conditioned tasks, we use a frozen pre-trained T5 language model to encode instructions into 512-dimensional embeddings, then project to token dimensions for cross-attention. 

\textbf{Proprioception Encoding.}
Proprioception is encoded through a 2-layer MLP.  We notice that progressively masking proprioceptive inputs with probability p during training helps alleviate overfitting to proprioception only and prevents the model from learning shortcuts that bypass visual understanding. This masking strategy can be important for dexterous manipulation tasks where robots might otherwise rely too heavily on proprioceptive feedback rather than developing robust visual-motor coordination, ultimately leading to more generalizable policies that can handle sensor noise and partial state observability in real-world deployment.

\textbf{Action Generation:} We predict action sequences of varying lengths depending on task complexity and use a 2-layer MLP to decode action tokens into continuous actions. We use action horizons of 4 steps for short-horizon simulation tasks (Adroit, DexArt, MetaWorld) and 16 steps for dexterous tasks in RoboTwin requiring bimanual coordination. For real-world tasks, we use 64 steps to account for execution delays and employ temporal ensembling to aggregate predicted actions over multiple timesteps, ensuring smoother temporal transitions and avoiding abrupt motion discontinuities for better stability and safety. We use an observation history of 2 timesteps for all tasks to provide temporal context while maintaining computational efficiency.

\textbf{Baseline Architecture.} We use the U-Net architecture as the diffusion network for both 2D and 3D diffusion/flow matching policies, following their original papers and code. While Diffusion Policy has both CNN and transformer variants available, we use the U-Net version as it demonstrates superior performance in our experiments.

\subsection{ManiFlow Training Details.} 

We employ a single-stage training approach that jointly optimizes flow matching and consistency objectives without requiring pre-trained teacher models. Rather than directly constraining velocities at intermediate points to be identical along the flow path, which often yields trivial solutions and unstable training, we learn mappings from any partially-noised data point to the final target data point, ensuring self-consistency throughout the ODE trajectory.
We provide the 
pseudocode for different times sampling strategies in Alg.~\ref{alg:time_sampling} and \ours training in Alg.~\ref{alg:maniflow_alg}.

\textbf{Joint Training Strategy.} To reduce the training cost of \ours, our training batch consists of two components with different batch ratios: 75\% for flow matching training and 25\% for consistency training. During flow matching training, we set $\Delta t = 0$ to predict instantaneous velocity at timestep $t$, while consistency training uses sampled $\Delta t$ from a continuous uniform distribution to enforce consistency across different points on the same trajectory. Additionally, we use different time sampling strategies for $t$: Beta distribution for flow matching to emphasize the high-noise regime, and discrete uniform sampling for consistency training to cover the full denoising trajectory.

\textbf{Target Time Conditioning.} A key design choice in our velocity prediction is the target timestep conditioning. We evaluate two modes: \textit{absolute} mode, where the model predicts velocity toward $t + \Delta t$, and \textit{relative} mode, where it predicts velocity for step size $\Delta t$. Empirically, we find that the relative mode ($\Delta t$ conditioning) achieves better performance than the absolute mode, as it provides more direct step-size information for the model to learn appropriate velocity magnitudes. 

\textbf{EMA Stabilization.} The exponential moving average (EMA) model plays a crucial role in stabilizing consistency training \cite{song2023consistency}. During consistency training, we require reliable velocity predictions at future timesteps to compute consistency targets, but using the current model (which is being updated) can lead to training instability due to rapidly changing predictions. Instead, we maintain an EMA version of the model parameters $\theta^- = \mu \theta^- + (1-\mu) \theta$, where $\mu$ is the momentum coefficient. This EMA model provides stable target generation for consistency training by offering slowly-evolving, more reliable velocity predictions at intermediate timesteps. The EMA mechanism ensures that consistency targets remain relatively stable across training iterations, preventing oscillations and enabling smooth convergence of the joint flow matching and consistency objectives.

\subsection{Failure Cases.}
We observe that ManiFlow fails in tasks that require detailed contact information and precise force feedback, such as delicate assembly operations or compliant insertion tasks. This limitation stems from ManiFlow's design focus on kinematic control rather than force-based interactions, lacking the tactile sensing and force control capabilities necessary for tasks where contact dynamics are critical for success. However, we believe incorporating tactile feedback as an additional modality would significantly enhance ManiFlow's capability to handle more contact-rich manipulation tasks and broaden its applicability.

\section{Simulation Experiments.} 

\subsection{Training Details.} 
We collect varying amounts of demonstrations across benchmarks based on task complexity: 10 demonstrations per task for Adroit and MetaWorld, 50 for RoboTwin, and 100 for DexArt. To ensure rigorous and fair evaluation, all models are trained and tested under identical conditions across multiple benchmarks. For the RoboTwin benchmark, models are trained for 2000 epochs, with performance evaluated on the final checkpoint over 100 episodes. For the Adroit and DexArt benchmarks, models are trained for 3000 epochs, with performance assessed every 50 epochs over 20 episodes. The final performance metric is computed as the average of the top five success rates to account for potential performance variations. 
In the MetaWorld benchmark, we specifically focus on the more challenging language-conditioned multi-task learning scenario rather than single-task evaluation. This decision stems from the observation that both baseline 3D diffusion policy and our method consistently achieve near-perfect success rates (approximately 90\% to 100\%) in single-task settings for most tasks, making it difficult to meaningfully differentiate their capabilities. The language-conditioned multi-task setting provides a more nuanced assessment of model performance.
For all benchmarks, we report both mean success rates and standard deviations across three independent training seeds to provide a comprehensive view of model performance and stability. This evaluation protocol, with consistent metrics and multiple seeds, ensures robust and reliable performance comparisons across all tested approaches.

\subsection{Simulation Benchmark}
MetaWorld contributes single-arm manipulation scenarios such as door opening and tool use, while Adroit specializes in dexterous manipulation using a shadow dexterous hand for precise finger control tasks like in-hand manipulation and pen twirling. DexArt introduces challenging dexterous tasks tested on unseen articulated objects, such as lifting a bucket and turning on a faucet with a revolute joint, while RoboTwin complements the suite with realistic simulation environments and a variety of bimanual dexterous manipulation tasks. This carefully curated benchmark selection enables a thorough evaluation of our policy's generalization capabilities across different environments, task complexities, and skill sets, providing comprehensive insight into its robustness and adaptability while maintaining direct relevance to real-world applications. The visualization of simulation tasks across these 4 benchmarks is shown in Fig.~\ref{fig:sim_tasks}.

\begin{figure*}[t]
    \centering
    \includegraphics[width=1.0\textwidth]{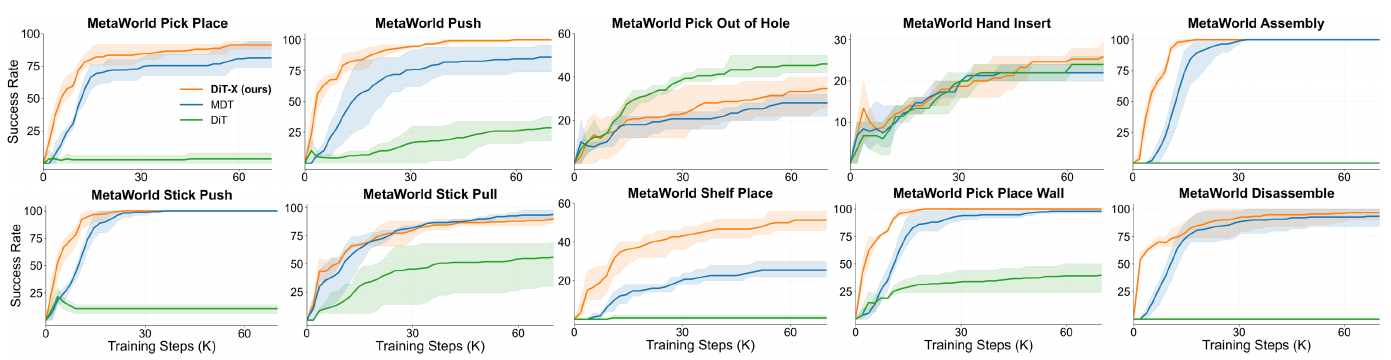}
    \vspace{-0.1in}
    \caption{\textbf{Comparison between DiT, MDT, and \ours's DiT-X block:} Language-conditioned multi-task learning curves for 10 MetaWorld hard tasks. DiT-X demonstrates faster convergence towards higher accuracy, highlighting superior multi-modal conditioning capabilities.}
    \label{fig:DiT_learning_efficiency}
\end{figure*}

\subsection{Ablation}

\begin{figure*}[t]
    \centering
    \includegraphics[width=1.0\textwidth]{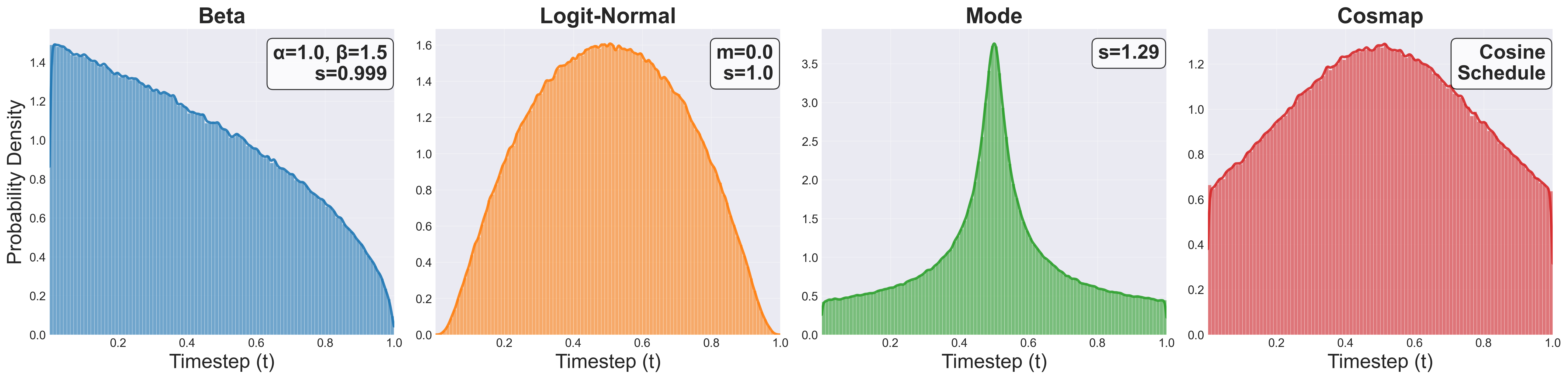}
    \caption{\textbf{Comparison of timestep sampling strategies for flow matching models}. We show the sample probability density of different timestep $t \in [0,1]$. The Beta distribution ($\alpha=1.0, \beta=1.5, s=0.999$) concentrates samples near t=0 (early noise levels), the logit-Normal distribution ($m=0.0, s=1.0$) provides balanced sampling around $t=0.5$, the Mode distribution ($s=1.29$) strongly favors midpoint during training through a scale parameter $s$, and Cosmap follows a cosine schedule. Histograms represent empirical sample frequencies, while smooth curves show estimated probability distributions. We provide pseudo code for each sampling strategy in Alg.~\ref{alg:time_sampling}.}
    \label{fig:time_distribution}
    \vspace{-0.1in}
\end{figure*}

\textbf{More Scaling Comparison.} We evaluate both 3D ManiFlow Policy and 3D Diffusion Policy across varying numbers of demonstrations on the Pick Apple Messy dexterous task from the RoboTwin benchmark, which requires picking apples from cluttered environments with distractors and random positions. 
As shown in Fig.~\ref{fig:maniflow_scaling}, ManiFlow exhibits strong scaling performance, increasing from 3.7\% with 10 demos to 57.3\% with 100 demos, and reaching 79.0\% with 500 demos using point cloud coordinates only, significantly outperforming the 3D diffusion policy baseline, which plateaus at 32.7\%. 
Adding RGB information further enhances performance, achieving 86.0\% at 100 demos and continuing to improve to 92.3\% at 500 demos.
This scaling capability stems from ManiFlow's more capable DiT-X architecture and efficient consistency training objective that better leverages more abundant data for learning complex dexterous behaviors. 
We expect \ours to achieve even better performance with larger, more diverse datasets.

\textbf{Ablation on Time Scheduler.} We ablate the scheduler choices of timestep $t$ and stepsize $\Delta t$  on 7 tasks from Adroit and MetaWorld benchmarks. For sampling $t$, as shown in Tab.~\ref{table: ablate time schedular}, while other schedulers like uniform, Cosmap, Mode, and especially logit-normal achieve reasonable results, the beta scheduler consistently outperforms them. The key advantage stems from its emphasis on lower timesteps with higher noise levels, which is particularly beneficial for robotic action prediction. This finding aligns with the insight that robot observations provide rich constraints on possible actions, making the learning of noise-conditioned policies especially important in the high-noise regime. 
For $\Delta t$ sampling, continuous time sampling shows better performance than discrete sampling.

\input{tables/ablation_time_schedular}

\input{tables/ablate_denoising_steps}

\textbf{Comparison Across Diffusion and Flow-Matching Training Objectives.} We evaluate \ours against representative generative models with different training objectives. Diffusion Policy \cite{chi2023diffusion} serves as our primary diffusion-based baseline given its strong performance in robotic control. For flow matching approaches, we include Rectified Flow \cite{liu2022flow}, which introduces a simplified training objective optimizing straight trajectories in latent space, Consistency-FM \cite{yang2024consistency}, which leverages velocity consistency to improve sample quality, and the shortcut model \cite{frans2024one}. which conditions on the additional step size and enforces self-consistency to improve generation quality. As shown in Tab.~\ref{table: compare with more baselines}, \ours consistently outperforms these baselines across diverse manipulation scenarios, demonstrating the effectiveness of our proposed training objective for robotic control tasks.

\input{tables/ablation_flow_baseline}

\textbf{ManiFlow as a Versatile and Effective Policy Head.} The broad applicability of ManiFlow is demonstrated through its successful integration into the established 3D Diffuser Actor \cite{ke20243d} (3D-DA) architecture as a policy head. As shown in Tab.~\ref{table:calvin}, single-step inference with ManiFlow (avg sequence 3.67) outperforms the original 25-step DDPM (avg sequence 3.35), achieving 25 times inference speedup. The performance advantage becomes more pronounced for longer instruction sequences, where our 10-step ManiFlow achieves a 0.68 higher average sequence length. Notably, the improvement is particularly significant for longer-horizon tasks, with ManiFlow showing substantial gains in completing 4-instruction (73.0\% vs 53.3\%) and 5-instruction chains (65.7\% vs 41.2\%). These promising results demonstrate ManiFlow's potential as an efficient and effective replacement for existing diffusion-based policy heads across robotic learning frameworks.


\input{tables/calvin}
\input{tables/results_dexterous}
\input{tables/results_metaworld}

\section{Real World Experiment}
\label{sec:appendix:real_world_exp_details}

\subsection{Real-World Setups}
\begin{figure*}[htbp]
    \centering
    \includegraphics[width=1.0\textwidth]{ 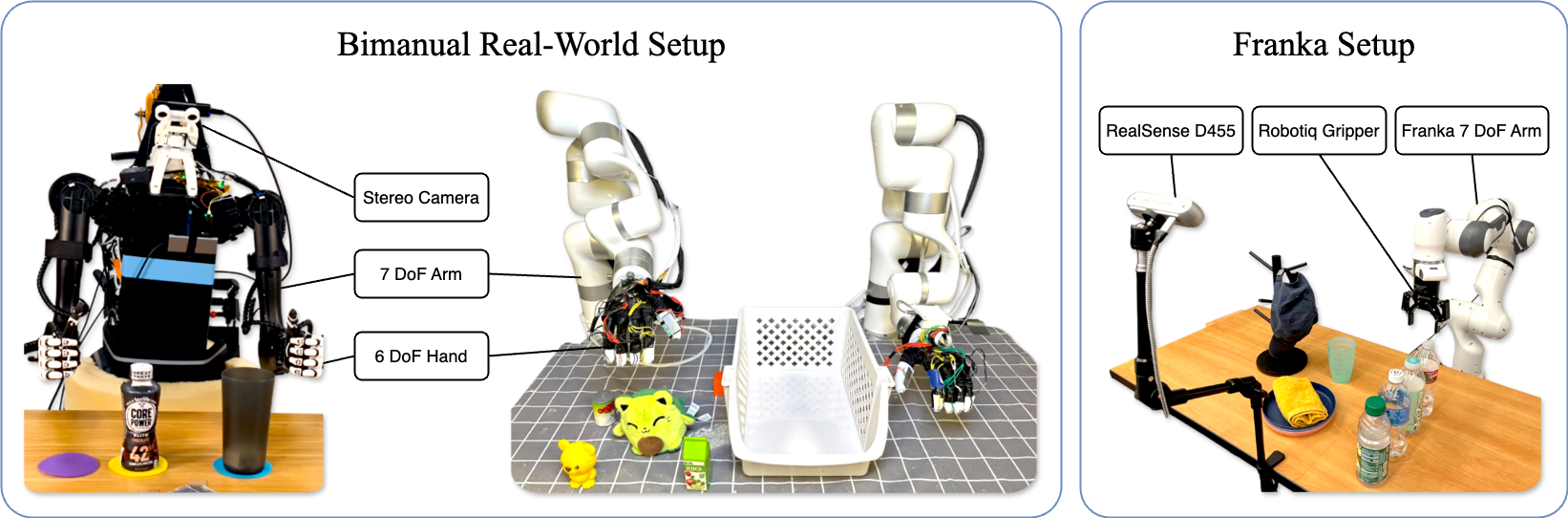}
    \caption{\textbf{Real-World Setup.} The experimental setup includes three configurations: (1) a bimanual Unitree H1 humanoid robot with 7-DoF arms, anthropomorphic hands, and a gimbal-mounted stereo camera; (2) a Bimanual 7DoF xArm setup with PSYONIC Ability Hands and an Intel RealSense L515 camera; and (3) a Franka Emika Panda robot with a Robotiq gripper and a statically positioned Intel RealSense D455 camera.}
    \label{fig:real_world_setup}
\end{figure*}

We evaluate ManiFlow’s performance on three distinct robot setups: the Unitree H1 humanoid robot, the bimanual xArm 7 robot configuration, and the Franka Emika Panda robot. Each setup is evaluated on a unique set of tasks designed to assess ManiFlow’s manipulation capabilities across diverse scenarios. Fig.~\ref{fig:real_world_setup} provides a visual overview of the experimental setups, including robot configurations and task environments. The details of each setup are as follows:
\begin{enumerate}[(a)]
    \item \textbf{Humanoid Setup.} The Unitree H1 is a full-sized humanoid robot equipped with two 7-DoF arms and anthropomorphic hands featuring 28-DoF (two 7-DoF arms + two 6-DoF anthropomorphic Inspire hands + 2-DoF active head). It is equipped with a gimbal-mounted ZED stereo camera, enabling active perception and spatial awareness. The humanoid's anthropomorphic hand design with 12 total DoF per hand (6 actuated, 6 underactuated through linkage mechanisms) requires sophisticated multi-finger coordination.
    
    \item \textbf{Bimanual Setup.} This setup consists of two UFACTORY xArm 7 robotic arms paired with two 6-DoF PSYONIC Ability Hands featuring 26-DoF in total, following the experiment configuration used in Bunny-VisionPro~\cite{ding2024bunny}. An Intel RealSense LiDAR L515 camera, positioned in front of the setup, provides visual observations.
    
    \item \textbf{Single-Arm Setup.} This configuration uses a 7-DoF Franka Emika Panda robot with a Robotiq parallel gripper. The robot is mounted statically, and an Intel RealSense D455 RGB-D camera provides external visual observations.
\end{enumerate}

\textbf{Humanoid vs. Bimanual Setup.}
The key differences include both perception and hardware complexity: 
\begin{enumerate}[(a)]
\item \textbf{Perception:} The Humanoid Setup uses first-person active sensing with a 2-DoF gimbal-mounted stereo camera that moves with the operator's head during data collection, requiring the policy to learn coordinated head-arm movements and handle dynamic viewpoints from training data. In contrast, the Bimanual Setup uses a static third-person view camera, providing a consistent but relatively limited perspective. 
The humanoid's active perception adds complexity as the policy must learn optimal head movements while managing visual instabilities from camera motion. 
\item \textbf{Hardware Complexity:} 
Humanoid setup present greater control challenges due to quasi-direct-drive motors that have gear clearance and reduced accuracy compared to precision industrial arms (UFactory xArms) used in bimanual setups. They also feature complex anthropomorphic kinematic chains with additional singularities and workspace limitations from human-like proportions. These mechanical imprecisions and kinematic constraints create significant challenges for policy learning in dexterous manipulation, as the learned policies must compensate for hardware inconsistencies and coordinate more complex joint configurations for fine-grained tasks.
\end{enumerate}

\subsection{Task Descriptions}
\begin{figure}[htbp]
    \centering
    \includegraphics[width=0.95\linewidth]{ 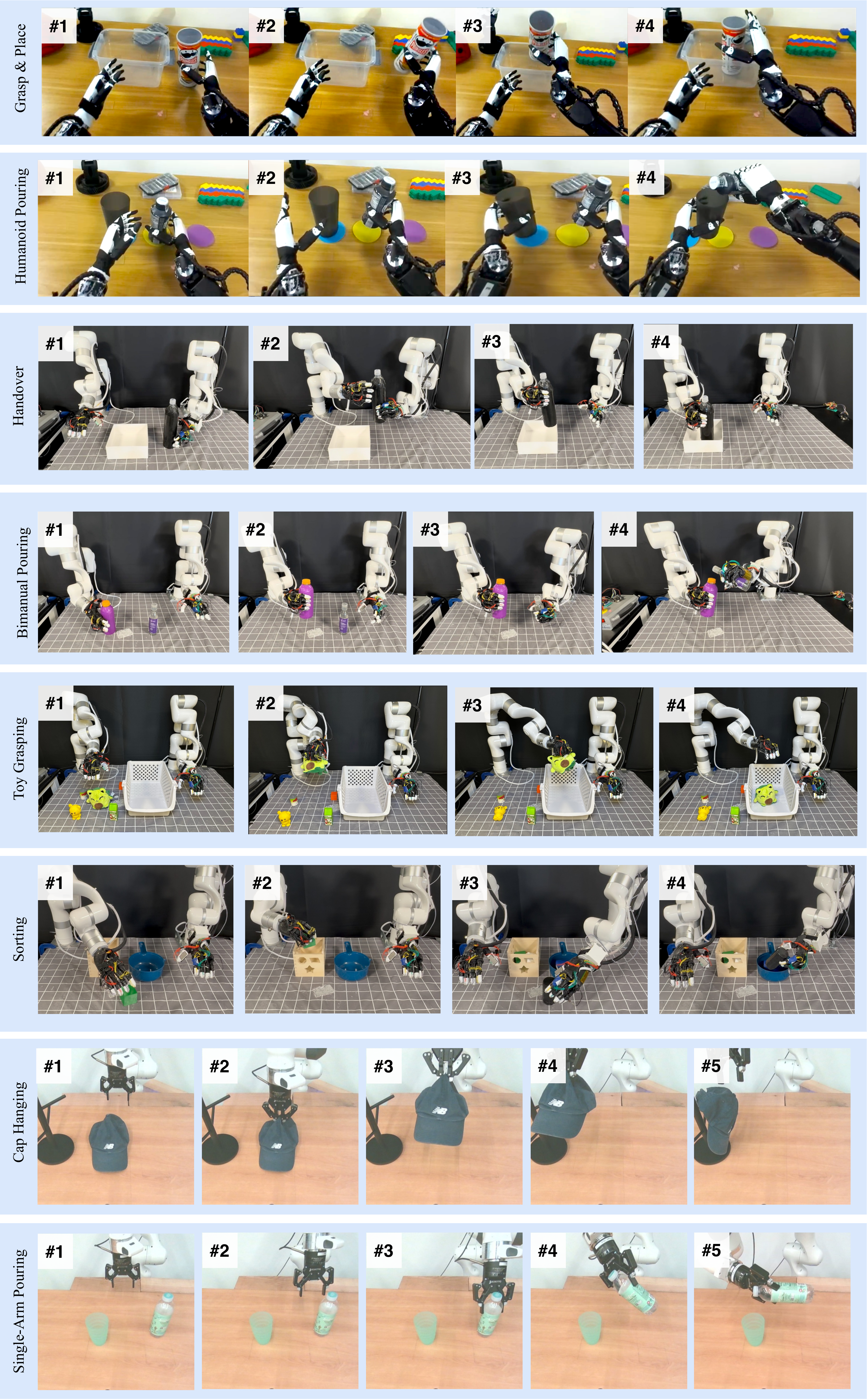}
    \caption{\textbf{Tasks Trajectories.} Illustration of the task trajectories, including Humanoid Grasp \& Place, Humanoid Pouring, Bimanual Handover, Bimanual Pouring, Bimanual Toy Grasping, Single-Arm Cap Hanging, and Single-Arm Pouring.}
    \label{fig:task_descript}
\end{figure}

We evaluate ManiFlow on eight real-world tasks, consisting of \textbf{(i)} two tasks evaluated on \textit{Humanoid Setup}: \textbf{Humanoid Grasp \& Place}, \textbf{Humanoid Pouring}, \textbf{(ii)} four tasks on \textit{Bimanual Setup}: \textbf{Bimanual Handover}, \textbf{Bimanual Pouring}, \textbf{Bimanual Toy Grasping}, \textbf{Bimanual Sorting}, and \textbf{(iii)} two tasks on \textit{Single-Arm Setup}: \textbf{Single-Arm Cap Hanging}, and \textbf{Single-Arm Pouring}. Notably, Bimanual Toy Grasping is a Single-Arm task executed within the bimanual setup. The first word of each task name specifies the corresponding real-world setup. Fig.~\ref{fig:task_descript} shows the trajectories of the tasks. Detailed task descriptions are provided below:

\begin{enumerate}[(a)] 
\item  \textbf{Humanoid Grasp \& Place}: The right hand grasps a water bottle and places it into a container. Evaluated on \textit{Humanoid Setup}.

\item \textbf{Humanoid Pouring}: The left hand grasps a cup and holds it. The right hand grasps a bottle and pours it into the cup accurately. Evaluated on \textit{Humanoid Setup}.

\item \textbf{Bimanual Handover}: The left hand grasps a bottle and hands it over to the right hand. The right hand then places the bottle into a box. Evaluated on \textit{Bimanual Setup}.

\item \textbf{Bimanual Pouring}: Both hands grasp separate water bottles. The left hand performs a pouring motion above the bottle held by the right hand. Evaluated on \textit{Bimanual Setup}.

\item \textbf{Bimanual Toy Grasping}: The right hand grasps a toy and places it into a basket, while randomly placed distractors interfere with the grasp. Evaluated on \textit{Bimanual Setup}.

\item \textbf{Bimanual Sorting}: Continuously sorts three objects, with the right hand placing cubes into a box and the left hand sorting cylinders into a pot. Evaluated on \textit{Bimanual Setup}.

\item \textbf{Single-Arm Cap Hanging}: The gripper grasps a cap and precisely positions it onto a hook. Evaluated on \textit{Single-Arm Setup}.

\item \textbf{Single-Arm Pouring}: The gripper grasps a bottle and pours it into a cup on the table. Evaluated on \textit{Single-Arm Setup}.
\end{enumerate}

\subsection{Data Collection}

\begin{enumerate}[(a)]
    \item \textbf{Humanoid Setup}: We follow the data collection approach outlined in Open-TeleVision~\cite{cheng2024tv}, using the Apple Vision Pro for teleoperation.
    \item \textbf{Bimanual Setup}: We adopt the same data collection methods as Bunny-VisionPro~\cite{ding2024bunny}, using Apple Vision Pro to teleoperate the bimanual hand-arm setup. Approximately 50 demonstrations are collected for each task.
    \item \textbf{Single-Arm Setup}: We use Oculus VR teleoperation, collecting 70--80 demonstrations per task. During data collection, objects are varied in type, location, and orientation to encourage generalization. 
\end{enumerate}

\subsection{Evaluation Metrics}
To assess generalization, we evaluate the model under the following categories:
\begin{itemize}
    \item \textbf{Seen Object:} Using objects and configurations from the training dataset.
    
    \item \textbf{Unseen Objects:} Using novel object types not present in training.

    \item \textbf{Perturbations:} Including Distractors in the Scene.

\end{itemize}

\begin{table}[htbp]
\centering
\vspace{-0.05in}

\resizebox{1.0\textwidth}{!}{%
\begin{tabular}{l|ccc}
\toprule
\textbf{Task} & \textbf{\# of Seen Objs} & \textbf{\# of Unseen Objs} & \textbf{\# of Eval Trials/Obj} \\
\midrule
Humanoid Grasp \& Place & 4 & 2 & 10 \\
Humanoid Pouring & 2 & 2 & 10 \\
Bimanual Handover & 3 & 2 & 10 \\
Bimanual Pouring & 4 & 2 & 10 \\
Bimanual Toy Grasping & 5 & 3 & 10 \\
Bimanual Sorting & 6 & 4 & 2.8 \\
Single-Arm Pour Water & 4 & 2 & 3.3 \\
Single-Arm Cap Hanging & 2 & 1 & 5 \\
\bottomrule
\end{tabular}}
\vspace{1ex}
\caption{\textbf{Number of Seen/Unseen objects for Each Task.} In Bimanual Toy Grasping, each trial involves a mixed set of objects, so we report the average number of trials per object. For other tasks, we record the exact number of trials per object.}
\vspace{-0.1in}
\label{tab:num_objects_each_task}
\end{table}


For both Seen and Unseen Object, we evaluate each object with the number of trials as shown in Tab. ~\ref{tab:num_objects_each_task}. 
Overall, except for Bimanual Sorting, we evaluated each method in 305 rollouts (29 objects with 10 trials each for the Humanoid and Bimanual settings and 15 trials total for the Single-Arm setting).

\subsection{Evaluation Details for Bimanual and Humanoid tasks} 

\begin{figure}[htbp]
    \centering
    \vspace{-0.15in}
    \includegraphics[width=0.85\linewidth]{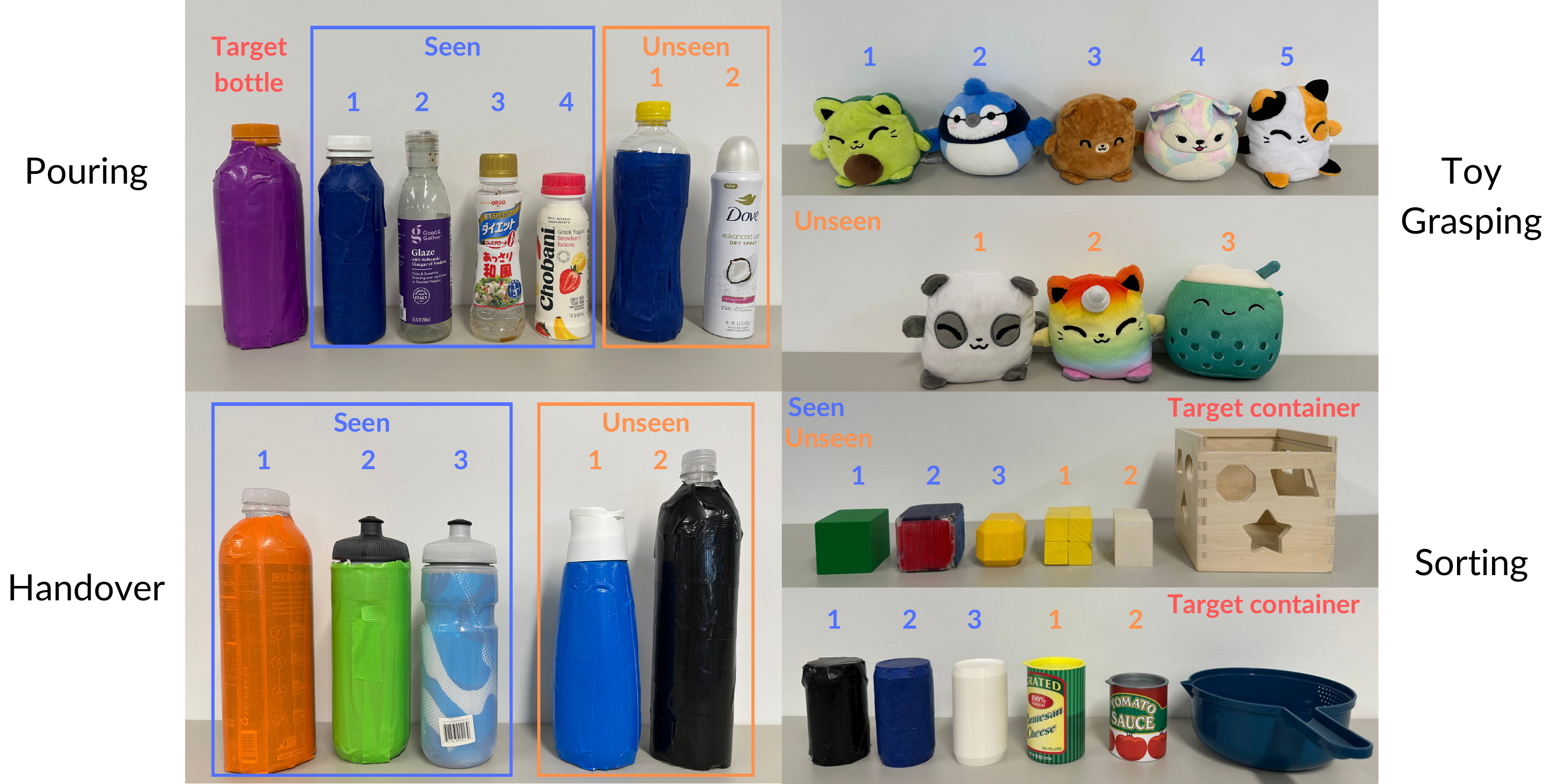}
    \caption{\textbf{Objects in Bimanual Setting.} The objects observed during the demonstration collection and the unseen objects are shown above. The objects selected represent a variety of geometries, with many differing in scale. (i) Pouring: The left hand grasps a seen or unseen bottle and performs a pouring motion above a target bottle held by the right hand. (ii) Handover: The left hand grasps a seen or unseen bottle and hands it over to the right hand. The right hand then places the bottle into a box. (iii) Toy Grasping: The right hand grasps a seen or unseen toy and places it into a basket. (iv) Sorting: The right hand sorts cubes and the left hand sorts cylinders into their respective containers.}
    \label{fig:bi_objects}
    \vspace{-0.15in}
\end{figure}

Fig.~\ref{fig:bi_objects} illustrates our sets of seen and unseen objects. (i) Bimanual Pouring involves one bottle serving as the target while another pours into it. The task demands precise grasping and rim alignment, so we choose bottles of varying sizes, shapes, and textures to evaluate the policy’s generalizability. (ii) Handover requires the robot to accurately grasp and transfer bottles. Thus, we select bottles of different shapes and sizes to assess performance. (iii) Toy Grasping primarily tests the policy’s spatial generalizability and its ability to operate amidst distractors. To this end, we select toys of similar sizes but diverse shapes. (iv) Sorting requires the policy to distinguish between the geometries of cubes and cylinders. We select cubes and cylinders with subtle differences in shape and scale.

\begin{figure}[htbp]
    \centering
    \includegraphics[width=\linewidth]{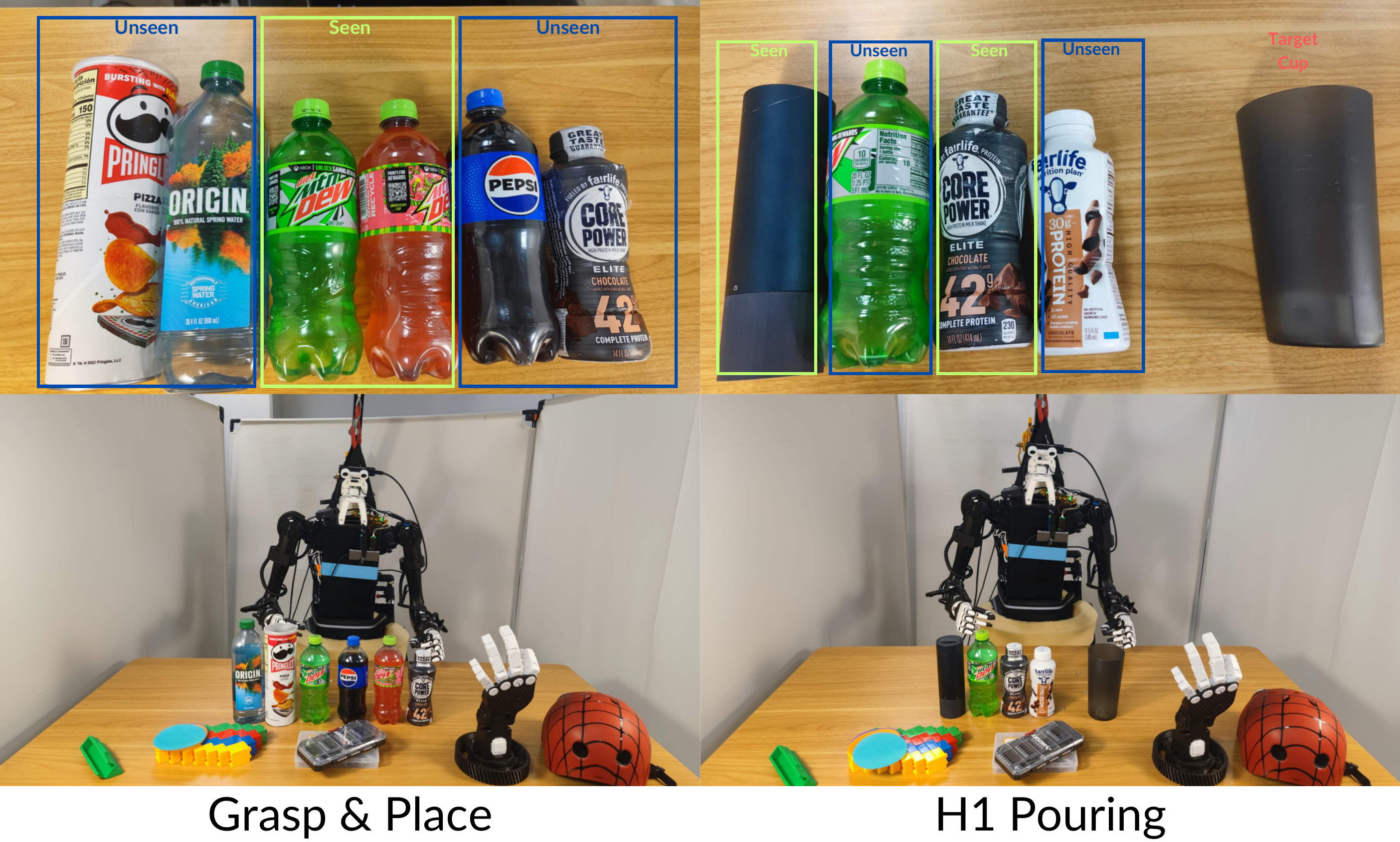}
    \caption{\textbf{Objects in Humanoid Setting.} The seen and unseen objects in the H1 setting are shown above, along with their relative sizes compared with H1 and the experiment environment.}
    \label{fig:h1_objects}
\end{figure}

Fig.~\ref{fig:h1_objects} shows our seen and unseen objects in Humanoid setting. Grasp \& Place requires accurately grasping the object and placing them into the basket, while Pouring requires the robot to grasp accurately the objects and align their poses well with the cup. We carefully select objects of varying shapes and scales, ensuring that the system encounters a wide range of object properties and tests its ability to handle different geometries and dimensions effectively.

\begin{figure}[htbp]
    \centering
    \includegraphics[width=\linewidth]{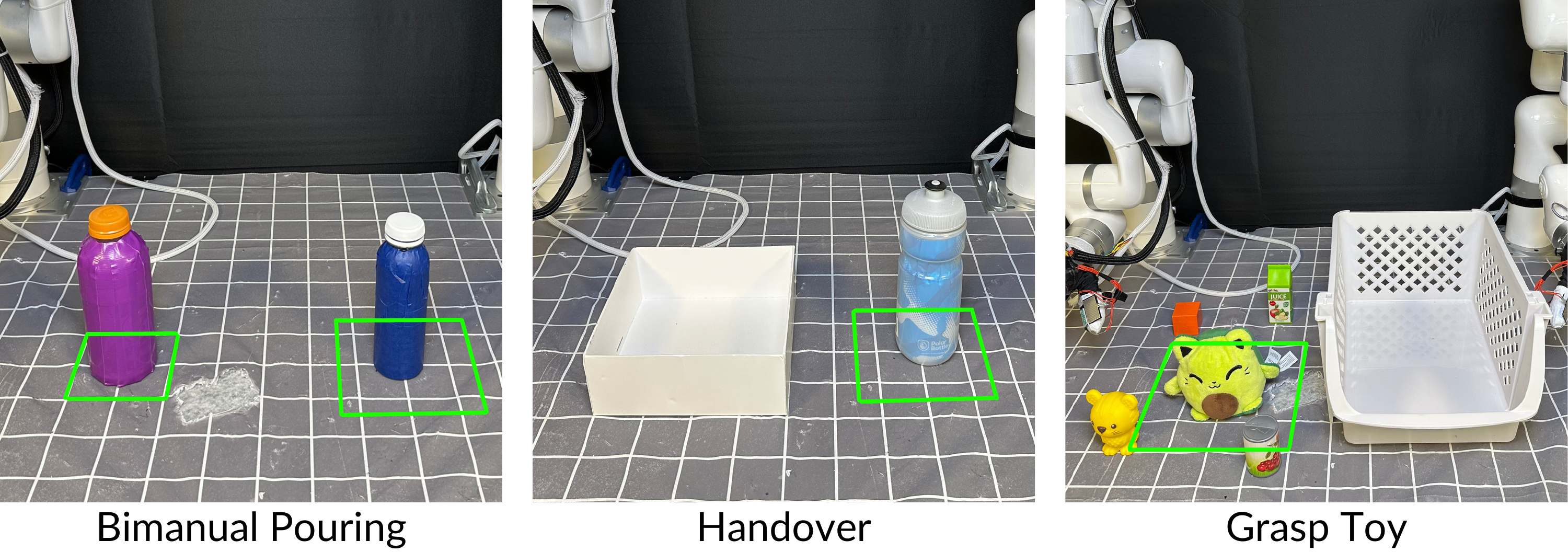}
    \caption{\textbf{Testing area of Bimanual Tasks.} The testing areas for our bimanual tasks are highlighted as green quadrilaterals. (i) Bimanual Pouring designates a 10.5cm × 10.5cm area for the target bottle and a 15cm × 15cm area for the pouring bottle. (ii) Handover positions the bottle within a 15cm × 15cm area, while the box may experience displacement perturbations of approximately 1.5cm in all directions. (iii) Toy Grasping places the toy within a 21cm × 21cm area, with distractors randomly arranged around it. Additionally, the basket may undergo front-back displacement perturbations of around 2.5cm in each direction.}
    \label{fig:bi_testing_area}
\end{figure}

\begin{figure}[htbp]
    \centering
    \includegraphics[width=\linewidth]{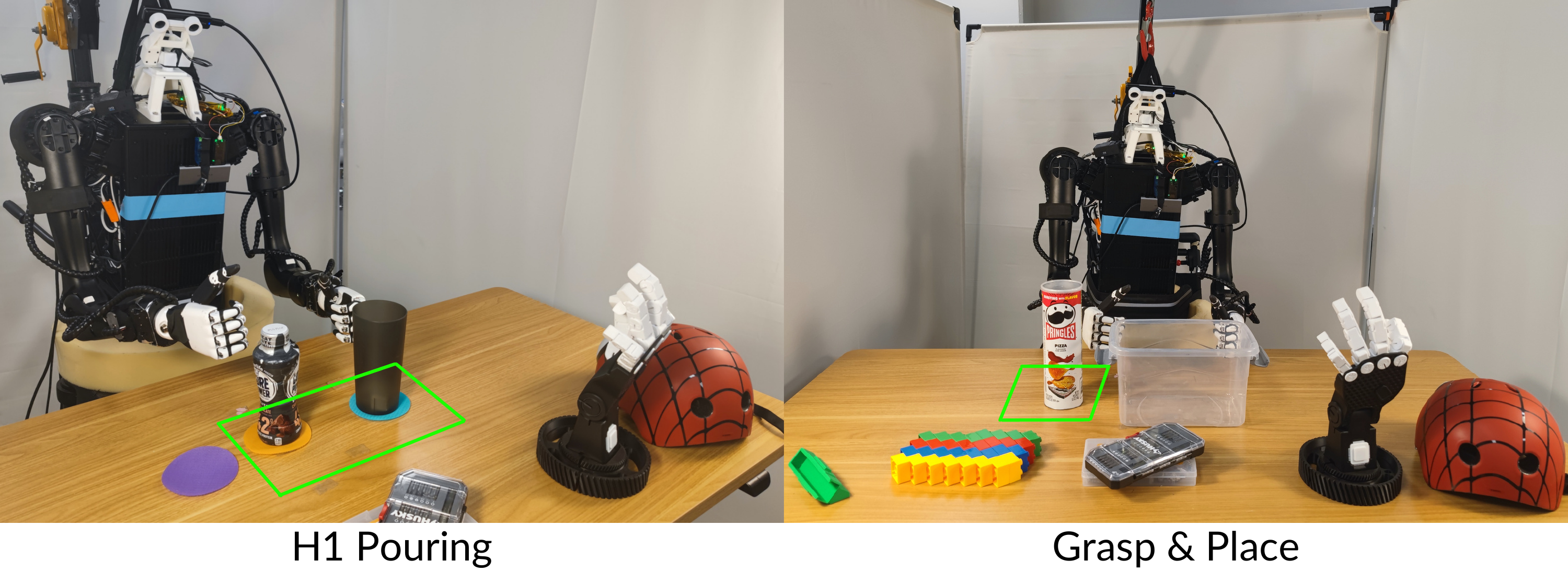}
    \caption{\textbf{Testing area of H1 Tasks.} The testing areas for our H1 tasks are highlighted as green quadrilaterals. (i) H1 Pouring positions the bottle and cup in front of H1, with variations in placement across different directions. (ii) Grasp \& Place situates the object to one side of H1, also allowing for positional variations in multiple directions.}
    \label{fig:h1_testing_area}
\end{figure}

\textbf{Single-Arm Setup Data Collection Details.} Specific data collection details for each task are provided below:

\begin{itemize}
    \item \textbf{Pouring Water:} The gripper grasps a water bottle and pours water into a cup.  Three types of water bottles are used. Bottle locations are randomized within a 10~cm $\times$ 20~cm area, and cup locations within an 8.5~cm $\times$ 20~cm area.
    \item \textbf{Cap Hanging:} Two types of caps are used. Cap locations are randomized within a 24~cm $\times$ 30~cm area, with orientations varying within 10--20 degrees.
\end{itemize}

\section{ManiFlow Training Algorithms}

\begin{algorithm}
\caption{Timestep Sampling Strategies for Flow Matching}
\label{alg:time_sampling}
\begin{algorithmic}[1]
\STATE \textbf{Beta Sampling} ($\alpha=1.0$, $\beta=1.5$, $s=0.999$):
\STATE \hspace{0.5cm} Sample $u \sim \text{Beta}(\alpha, \beta)$
\STATE \hspace{0.5cm} $t \leftarrow s \cdot u$
\STATE 
\STATE \textbf{Logit-Normal Sampling} ($m=0.0$, $s=1.0$):
\STATE \hspace{0.5cm} Sample $z \sim \mathcal{N}(m, s^2)$
\STATE \hspace{0.5cm} $t \leftarrow \frac{1}{1 + e^{-z}}$
\STATE 
\STATE \textbf{Mode Sampling} ($s=1.29$):
\STATE \hspace{0.5cm} Sample $u \sim \text{Uniform}(0, 1)$
\STATE \hspace{0.5cm} $t \leftarrow 1 - u - s \cdot \left(\cos^2\left(\frac{\pi u}{2}\right) - 1 + u\right)$
\STATE \hspace{0.5cm} $t \leftarrow \max(0, \min(1, t))$
\STATE 
\STATE \textbf{Cosmap Sampling}:
\STATE \hspace{0.5cm} Sample $u \sim \text{Uniform}(0, 1)$
\STATE \hspace{0.5cm} $t \leftarrow 1 - \frac{1}{\tan\left(\frac{\pi u}{2}\right) + 1}$
\STATE \hspace{0.5cm} $t \leftarrow \max(0, \min(1, t))$
\end{algorithmic}
\end{algorithm}

\begin{algorithm}[htbp]
\caption{ManiFlow Model Training}
\label{alg:maniflow_alg}
\begin{algorithmic}[1]
\WHILE{not converged}
    \STATE \textbf{Sample Data Points:}
    \STATE \hspace{1em} $x_0 \sim \mathcal{N}(0, I)$ \COMMENT{Sample from noise distribution}
    \STATE \hspace{1em} $x_1 \sim D$ \COMMENT{Sample from data distribution}
    \STATE \hspace{1em} \textbf{if} Flow Matching Training \textbf{then}
    \STATE \hspace{2em} $t \sim \text{Beta}(\alpha, \beta)$ \COMMENT{Sample time from Beta distribution}
    \STATE \hspace{2em} $\Delta t \leftarrow 0$ \COMMENT{No time step size for Flow matching training}
    \STATE \hspace{1em} \textbf{else if} Consistency Training \textbf{then}
    \STATE \hspace{2em} $t \sim \mathcal{U}\{0, \frac{1}{T}, \frac{2}{T}, \ldots, \frac{T-1}{T}\}$ \COMMENT{Sample time from discretized [0,1) interval}
    \STATE \hspace{2em} $\Delta t,\Delta t'  \sim \mathcal{U}[0,1]$ \COMMENT{Sample time interval from uniform distribution}
    
    \STATE \textbf{Construct Linear Interpolation Path:}
    \STATE \hspace{1em} $x_t \leftarrow (1-t) x_0 + t x_1$ \COMMENT{Current interpolated point}
    \STATE \hspace{1em} \textbf{if} Consistency Training
    \textbf{then}
    \STATE \hspace{2em} $t_1 \leftarrow t + \Delta t$ \COMMENT{Next time step}
    \STATE \hspace{2em} $x_{t_1} \leftarrow (1 - t_1) x_0 + t_1 x_1$ \COMMENT{Next interpolated point}
    
    \STATE \textbf{Compute Target Velocities:}
    \STATE \hspace{1em} \textbf{if} Flow Matching Training \textbf{then}
    \STATE \hspace{2em} $v_{\text{target}} \leftarrow x_1 - x_0$ \COMMENT{Direct flow target}
    
    \STATE \hspace{1em} \textbf{else if} Consistency Training \textbf{then}
    \STATE \hspace{2em} $v_{t_1} \leftarrow v_{\theta^-}(x_{t_1}, t_1, \Delta t')$ \COMMENT{Velocity from flow EMA model}
    \STATE \hspace{2em} $\tilde{x}_1 \leftarrow x_{t_1} + v_{t_1} \cdot (1 - t_1)$ \COMMENT{ODE integration step}
    \STATE \hspace{2em} $v_{\text{target}} \leftarrow (\tilde{x}_1 - x_t) / (1-t)$ \COMMENT{Average velocity as consistency target}
    
    \STATE \textbf{Update Parameters:}
    \STATE \hspace{1em} $\mathcal{L} \leftarrow \|v_\theta(x_t, t, \Delta t) - v_{\text{target}}\|^2$ \COMMENT{Compute loss}
    \STATE \hspace{1em} $\theta \leftarrow \theta - \alpha \nabla_\theta \mathcal{L}$ \COMMENT{Gradient update}
    
\ENDWHILE
\end{algorithmic}
\end{algorithm}

%% file: tables/ablation_time_schedular.tex









\begin{table}[t]
\centering
\caption{\textbf{Ablation on Time Scheduler.} We compare different time schedulers of timestep $t$ for flow matching and stepsize $\Delta t$ for consistency training with our \ours policy.}
\label{table: ablate time schedular}
\resizebox{1.0\textwidth}{!}{%
\begin{tabular}{c|c|ccccccc|ccc}
\toprule
\centering Time & \centering Time Scheduler & Door & Pen & shelf-place & pick-place-wall & stick-pull & stick-push & disassemble & \textbf{Average} \\
\midrule
\multirow{5}{*}{$t$} & \textbf{Beta} & \ddbf{80.3}{1.2} & \ddbf{55.5}{5.8} & \dd{44.0}{9.1} & \ddbf{95.3}{0.9} & \dd{90.7}{0.9} & \ddbf{100.0}{0.0} & \ddbf{80.0}{1.6} & \ddbf{78.0}{2.8}\\
& Uniform & \dd{77.7}{0.9} & \dd{55.0}{2.9} & \dd{40.0}{4.3} & \dd{94.7}{5.0} & \dd{87.3}{4.7} & \ddbf{100.0}{0.0} & \dd{80.0}{4.3} & \dd{76.4}{3.2}\\
& Lognorm & \dd{79.5}{2.0} & \dd{55.0}{2.8} & \dd{43.3}{8.4} & \dd{94.7}{2.5} & \dd{90.7}{0.9} & \ddbf{100.0}{0.0} & \ddbf{81.0}{3.0} & \dd{77.7}{2.8}\\
& Cosmap & \dd{80.3}{2.9} & \dd{52.0}{2.5} & \dd{44.0}{5.9} & \dd{93.3}{5.0} & \dd{88.0}{2.8} & \ddbf{100.0}{0.0} & \dd{82.0}{5.9} & \dd{77.1}{3.6}\\
& Mode & \dd{78.8}{5.9} & \dd{53.0}{2.7} & \dd{35.3}{2.5} & \dd{94.7}{6.2} & \dd{89.3}{3.4} & \ddbf{100.0}{0.0} & \dd{82.0}{3.3} & \dd{76.2}{3.4}\\
\midrule

\multirow{2}{*}{$\Delta t$}
& \textbf{continuous} & \ddbf{80.3}{1.2} & \ddbf{55.5}{5.8} & \ddbf{44.0}{9.1} & \ddbf{95.3}{0.9} & \ddbf{90.7}{0.9} & \ddbf{100.0}{0.0} & \dd{80.0}{1.6} & \ddbf{78.0}{2.8}\\

& discrete & \dd{78.7}{2.0} & \dd{52.0}{3.9} & \dd{37.3}{10.6} & \ddbf{95.3}{6.6} & \dd{90.0}{4.0} & \ddbf{100.0}{0.0} & \ddbf{80.7}{2.5} & \dd{76.3}{4.2}\\
\bottomrule
\end{tabular}}
\vspace{-0.1in}
\end{table}

%% file: tables/ablate_denoising_steps.tex
\begin{table}[t]
    \centering
    \caption{\textbf{Few-step Inference.}
    \ours achieves better efficiency with only a few inference steps compared to 3D Diffusion and Flow Matching Policy across 5 bimanual dexterous tasks on the RoboTwin benchmark.}
    \vspace{0.05in}\label{tab:ablate_manipulation_tasks}
    \resizebox{1.0\textwidth}{!}{
    \begin{tabular}{c|c|cccccc}
        \toprule
        Algorithm & Inference Step & \begin{tabular}[c]{@{}c@{}}Pick\end{tabular} & \begin{tabular}[c]{@{}c@{}}Diverse\end{tabular} & \begin{tabular}[c]{@{}c@{}}Dual\end{tabular} & \begin{tabular}[c]{@{}c@{}}Empty\end{tabular} & \begin{tabular}[c]{@{}c@{}}Shoe\end{tabular} & Average \\
        \midrule
        3D Diffusion Policy & 10 & \dd{9.3}{3.7} & \dd{38.3}{7.1} & \dd{46.3}{2.5} & \dd{73.0}{0.8} & \dd{46.5}{2.5} & \dd{42.7}{3.3} \\
        \midrule
        3D Flow Matching Policy* & 10 & \dd{16.0}{7.1} & \dd{56.3}{6.6} & \dd{46.5}{0.5} & \dd{82.3}{1.7} & \dd{39.3}{15.5} & \dd{48.1}{6.3} \\
        \midrule
        \multirow{5}{*}{\textbf{3D ManiFlow Policy}} & 1 & \dd{42.7}{1.9} & \dd{75.3}{1.7} & \dd{53.7}{0.5} & \ddbf{83.0}{0.0} & \dd{63.7}{2.6} & \dd{63.7}{2.2} \\
        & 2 & \ddbf{43.3}{2.1} & \ddbf{76.3}{1.7} & \dd{54.0}{1.6} & \dd{82.0}{1.4} & \dd{66.7}{2.9} & \ddbf{64.5}{1.9} \\
        & 4 & \dd{38.3}{1.2} & \dd{72.7}{1.9} & \ddbf{54.3}{1.9} & \dd{75.3}{2.4} & \dd{67.3}{4.9} & \dd{61.6}{2.5} \\
        & 8 & \dd{41.3}{0.5} & \dd{72.7}{2.5} & \dd{53.7}{1.7} & \dd{72.3}{3.3} & \ddbf{68.3}{2.9} & \dd{61.7}{2.2} \\
        & 10 & \dd{42.0}{0.8} & \dd{72.3}{1.7} & \dd{54.0}{2.2} & \dd{72.7}{4.8} & \ddbf{68.3}{2.9} & \dd{61.9}{2.5} \\
        \bottomrule
    \end{tabular}
    }
\label{table: inference}
\end{table}

%% file: tables/ablation_flow_baseline.tex









\begin{table*}[h]
\centering
\caption{\textbf{Ablation on more generative models.} We include more diffusion and flow matching baselines for comparison. All variants use the same encoder and DiT-X architecture. }
\vspace{-0.05in}
\label{table: compare with more baselines}
\resizebox{1.0\textwidth}{!}{%
\begin{tabular}{l|ccccccc|cc}
\toprule

Algorithm $\backslash$ Task & Door & Pen & shelf-place &  pick-place-wall  & stick-pull & stick-push & disassemble & \textbf{Average} \\
\midrule
\textbf{\ours} & \ddbf{80.3}{1.2} & \ddbf{55.5}{5.8} & \dd{44.0}{9.1} & \ddbf{95.3}{0.9} & \dd{90.7}{0.9} & \ddbf{100.0}{0.0} & \ddbf{80.0}{1.6} & \ddbf{78.0}{2.8}\\
DDIM \cite{song2020denoising} & \dd{79.3}{5.2} & \dd{53.8}{1.0} & \dd{44.0}{5.7} & \dd{90.7}{7.4} & \ddbf{94.0}{1.6} & \ddbf{100.0}{0.0} & \dd{78.7}{1.9} & \dd{77.2}{3.3}\\
Rectified Flow \cite{liu2022flow} & \dd{78.2}{3.6} & \dd{49.7}{5.0} & \ddbf{46.0}{4.9} & \dd{88.7}{9.8} & \dd{88.0}{4.3} & \ddbf{100.0}{0.0}& \dd{79.3}{1.9} & \dd{75.7}{4.2}\\
Consistency-FM \cite{yang2024consistency} & \dd{79.7}{1.9} & \dd{52.2}{1.9} & \dd{42.0}{5.9} & \dd{92.0}{8.5} & \dd{88.7}{5.2} & \ddbf{100.0}{0.0} & \dd{79.3}{3.4} & \dd{76.3}{3.8}\\
Shortcut Model \cite{frans2024one} & \dd{80.0}{1.4} & \dd{52.2}{5.7} & \dd{40.7}{5.2} & \dd{93.3}{5.7} & \dd{89.3}{4.1} & \ddbf{100.0}{0.0} & \dd{78.0}{2.8} & \dd{76.2}{3.6}\\


\bottomrule
\end{tabular}}
\end{table*}

%% file: tables/calvin.tex


\newcommand{\tb}[3]{\setlength{\tabcolsep}{#2mm}\resizebox{0.98\textwidth}{!}{\begin{tabular}{#1}#3\end{tabular}}}
\begin{table}[H] 
    \centering
    \begin{adjustbox}{width=0.98\textwidth} 
    \tb{@{}l|cccccc@{}}{2.0}{
    & \multicolumn{6}{c}{Instruction completed in a row (1000 chains)} \\
    & 1 & 2 & 3 & 4 & 5 & Avg. Len \\
    \midrule
    RoboFlamingo~\cite{li2023vision} & 82.4 & 61.9 & 46.6 & 33.1 & 23.5 & 2.48 \\
    SuSIE~\cite{black2023zero} & 87.0 & 69.0 & 49.0 & 38.0 & 26.0 & 2.69 \\
    GR-1~\cite{wu2023unleashing} & 85.4 & 71.2 & 59.6 & 49.7 & 40.1 & 3.06 \\
    3D-DA (DDPM 25 steps) & 93.8$_{\pm 0.01}$ & 80.3$_{\pm 0.0}$ & 66.2$_{\pm 0.01}$ & 53.3$_{\pm 0.02}$ & 41.2$_{\pm 0.01}$ & 3.35$_{\pm 0.04}$ \\
    3D-DA (\ours 1-step) & 92.7$_{\pm 0.6}$ & 82.4$_{\pm 1.5}$ & 72.0$_{\pm 3.5}$ & 64.4$_{\pm 3.5}$ & 55.9$_{\pm 4.8}$ & 3.67$_{\pm 0.13}$ \\
    3D-DA (\ours 10-step) & $\mathbf{95.1}_{\pm 0.3}$ & $\mathbf{88.0}_{\pm 1.3}$ & $\mathbf{81.0}_{\pm 1.7}$ & $\mathbf{73.0}_{\pm 3}$ & $\mathbf{65.7}_{\pm 3.2}$ & $\mathbf{4.03}_{\pm 0.09}$ \\
    \bottomrule
    }
    \end{adjustbox}
    \vspace{1ex}
    \caption{\textbf{Zero-shot long-horizon evaluation on CALVIN} on 3 random seeds. 3D-DA \cite{ke20243d} with ManiFlow policy head achieves superior performance with fewer inference steps, especially for longer instruction sequences.}
    \label{table:calvin}
\end{table}

%% file: tables/results_dexterous.tex
\begin{table*}[t]
\centering
\caption{\textbf{Main results on 3 dexterous manipulation benchmarks.}}
\vspace{-0.1in}
\label{table: simulation results}
\begin{flushleft}
\resizebox{1.0\textwidth}{!}{%
\begin{tabular}{l|c|ccc|c|cccc|c}
\toprule
\multirow{2}{*}{Algorithm $\backslash$ Task} & \multirow{2}{*}{Obs.} & \multicolumn{4}{c|}{\textbf{Adroit (10 demos)}} & \multicolumn{5}{c}{\textbf{DexArt (100 demos)}} \\
\cmidrule(lr){3-6} \cmidrule(lr){7-11}
 & & hammer & door & pen & \textbf{Average} & laptop & faucet & bucket & toilet & \textbf{Average} \\
\midrule
Diffusion Policy & Img & \dd{54.0}{3.6} & \dd{41.8}{2.7} & \dd{18.5}{2.5} & \dd{38.1}{2.9} 
& \dd{81.7}{2.1} & \dd{29.3}{2.1} & \dd{26.0}{2.4} & \dd{77.3}{1.9} & \dd{53.6}{2.1} \\
Flow Matching Policy & Img & \dd{55.7}{4.2} & \dd{40.0}{1.6} & \dd{21.2}{0.8} & \dd{39.0}{2.2} 
& \dd{81.7}{2.5} & \dd{31.3}{3.7} & \dd{24.0}{2.2} & \dd{76.3}{1.2} & \dd{53.3}{2.4} \\
\textbf{2D ManiFlow} & Img & \ddbf{100.0}{0.0} & \ddbf{67.0}{2.2} & \ddbf{56.0}{3.6} & \ddbf{74.3}{1.9} 
& \ddbf{85.7}{2.1} & \ddbf{32.3}{0.5} & \ddbf{29.7}{3.4} & \ddbf{77.7}{3.3} & \ddbf{56.3}{2.3} \\
\midrule  
3D Diffusion Policy & PC & \dd{100.0}{0.0} & \dd{76.7}{4.7} & \ddbf{56.7}{2.6} & \dd{77.8}{2.4} 
& \dd{89.7}{0.9} & \dd{41.7}{0.5} & \dd{31.3}{0.5} & \ddbf{79.7}{0.9} & \dd{60.6}{0.7} \\
3D Flow Matching* & PC & \dd{100.0}{0.0} & \dd{77.7}{6.1} & \dd{53.5}{3.9} & \dd{77.1}{3.3} 
& \dd{92.7}{1.2} & \dd{42.0}{0.8} & \dd{32.3}{1.9} & \ddbf{79.7}{0.5} & \dd{61.7}{1.1} \\
\textbf{3D ManiFlow} & PC & \ddbf{100.0}{0.0} & \ddbf{80.3}{1.2} & \dd{55.5}{5.8} & \ddbf{78.6}{2.3} 
& \ddbf{93.0}{1.6} & \ddbf{45.0}{3.6} & \ddbf{35.3}{2.1} & \dd{79.3}{3.3} & \ddbf{63.2}{2.7} \\
\bottomrule
\end{tabular}}

\resizebox{1.0\textwidth}{!}{%
\begin{tabular}{l|c|ccccc|c|c}
\toprule
\multirow{2}{*}{Algorithm $\backslash$ Task} & \multirow{2}{*}{Obs.} & \multicolumn{6}{c|}{\textbf{RoboTwin (50 demos)}} & \multirow{2}{*}{\textbf{Overall Avg.}} \\
\cmidrule(lr){3-8}
 & & Pick Apple Messy & Diverse Bottles Pick & Dual Bottles Pick Hard & Empty Cup Place & Shoe Place & \textbf{Average} & \\
\midrule
Diffusion Policy & Img & \dd{17.0}{0.8} & \dd{36.3}{2.4} & \dd{41.3}{3.7} & \dd{42.0}{1.6} & \dd{7.3}{2.9} & \dd{28.8}{2.3} & \dd{39.4}{2.3} \\
Flow Matching Policy & Img & \dd{15.3}{1.9} & \dd{32.0}{4.5} & \dd{43.0}{0.0} & \dd{38.0}{5.4} & \dd{7.3}{1.7} & \dd{27.1}{2.7} & \dd{38.8}{2.5} \\
\textbf{2D ManiFlow} & Img & \ddbf{37.3}{4.8} & \ddbf{37.0}{1.6} & \ddbf{47.3}{2.1} & \ddbf{63.7}{1.2} & \ddbf{45.3}{3.7} & \ddbf{46.1}{2.7} & \ddbf{56.5}{2.4} \\
\midrule  
3D Diffusion Policy & PC & \dd{9.3}{3.7} & \dd{38.3}{7.1} & \dd{46.3}{2.5} & \dd{73.0}{0.8} & \dd{46.5}{2.5} & \dd{42.7}{3.3} & \dd{57.4}{2.2} \\
3D Flow Matching*& PC & \dd{16.0}{7.1} & \dd{56.3}{6.6} & \dd{46.5}{0.5} & \ddbf{82.3}{1.7} & \dd{39.3}{15.5} & \dd{48.1}{6.3} & \dd{59.9}{2.8} \\
\textbf{3D ManiFlow} & PC & \ddbf{42.0}{0.8} & \ddbf{72.3}{1.7} & \ddbf{54.0}{2.2} & \dd{72.7}{4.8} & \ddbf{68.3}{2.9} & \ddbf{61.9}{2.5} & \ddbf{66.5}{2.5} \\
\bottomrule
\end{tabular}}
\end{flushleft}
\vspace{-0.1in}
\end{table*}

%% file: tables/results_metaworld.tex
\begin{table*}[h]
\centering
\caption{\textbf{Language-conditioned Multi-task results on 48 Meta-World simulation tasks.} Results for using 10 demonstrations for each task are provided in this table.}
\vspace{-0.1in}
\label{table: simulation results}
\begin{flushleft}
\resizebox{1.0\textwidth}{!}{%
\begin{tabular}{l|ccccccc}
\toprule
& \multicolumn{7}{c}{\textbf{Meta-World (Easy)}} \\

 Alg $\backslash$ Task & Button Press & Button Press Topdown & Button Press Topdown Wall & Button Press Wall & Coffee Button & Dial Turn & Door Close \\
\midrule

 3D Diffusion Policy & \dd{62}{15} & \ddbf{100}{0} & \ddbf{100}{0} & \dd{72}{25} & \dd{73}{34} & \dd{53}{15} & \ddbf{100}{0} \\
 3D Flow Matching* & \dd{0}{0} & \ddbf{100}{0} & \ddbf{100}{0} & \dd{67}{31} & \dd{97}{5} & \ddbf{70}{7} & \ddbf{100}{0} \\
\textbf{3D \ours} & \ddbf{100}{0} & \ddbf{100}{0} & \ddbf{100}{0} & \ddbf{100}{0} & \ddbf{100}{0} & \dd{67}{13} & \ddbf{100}{0} \\

\bottomrule
\end{tabular}}

\resizebox{1.0\textwidth}{!}{%
\begin{tabular}{l|ccccccc}
\toprule
& \multicolumn{7}{c}{\textbf{Meta-World (Easy)}} \\

 Alg $\backslash$ Task & Door Lock & Door Open & Door Unlock & Drawer Close & Drawer Open & Faucet Close & Faucet Open \\
\midrule
 3D Diffusion Policy & \dd{0}{0} & \ddbf{100}{0} & \dd{98}{2} & \dd{88}{13} & \dd{98}{2} & \dd{92}{8} & \dd{83}{12} \\
 3D Flow Matching* & \dd{0}{0} & \ddbf{100}{0} & \ddbf{100}{0} & \dd{5}{7} & \ddbf{100}{0} & \dd{92}{6} & \ddbf{100}{0} \\
\textbf{3D \ours} & \ddbf{78}{14} & \ddbf{100}{0} & \ddbf{100}{0} & \ddbf{100}{0} & \ddbf{100}{0} & \ddbf{100}{0} & \ddbf{100}{0} \\

\bottomrule
\end{tabular}}

\resizebox{1.0\textwidth}{!}{%
\begin{tabular}{l|ccccccc}
\toprule
& \multicolumn{7}{c}{\textbf{Meta-World (Easy)}} \\

 Alg $\backslash$ Task & Handle Press & Handle Pull & Handle Pull Side & Lever Pull & Plate Slide & Plate Slide Back & Plate Slide Back Side \\
\midrule

3D Diffusion Policy & \ddbf{100}{0} & \dd{22}{17} & \dd{43}{6} & \dd{60}{12} & \dd{20}{18} & \dd{92}{12} & \ddbf{100}{0} \\
3D Flow Matching* & \ddbf{100}{0} & \dd{15}{18} & \dd{20}{7} & \dd{45}{11} & \dd{0}{0} & \dd{88}{10} & \ddbf{100}{0} \\
\textbf{3D \ours} & \ddbf{100}{0} & \ddbf{42}{10} & \ddbf{65}{7} & \ddbf{63}{19} & \ddbf{100}{0} & \ddbf{93}{9} & \ddbf{100}{0} \\

\bottomrule
\end{tabular}}

\resizebox{1.0\textwidth}{!}{%
\begin{tabular}{l|ccccccc}
\toprule
& \multicolumn{3}{c|}{\textbf{Meta-World (Easy)}} & \multicolumn{4}{c}{\textbf{Meta-World (Medium)}} \\

 Alg $\backslash$ Task & Plate Slide Side & Reach & Reach Wall & Window Close & Window Open & Basketball & Bin Picking \\
\midrule

 3D Diffusion Policy & \dd{92}{6} & \dd{48}{2} & \dd{25}{4} & \ddbf{100}{0} & \dd{83}{17} & \ddbf{100}{0} & \dd{0}{0} \\
 3D Flow Matching* & \dd{82}{14} & \dd{57}{10} & \dd{35}{11} & \ddbf{100}{0} & \dd{92}{6} & \dd{90}{4} & \dd{18}{6} \\
\textbf{3D \ours} & \ddbf{100}{0} & \ddbf{58}{19} & \ddbf{67}{9} & \ddbf{100}{0} & \ddbf{97}{2} & \ddbf{100}{0} & \ddbf{33}{2} \\

\bottomrule
\end{tabular}}

\resizebox{1.0\textwidth}{!}{%
\begin{tabular}{l|ccccccc}
\toprule
& \multicolumn{7}{c}{\textbf{Meta-World (Medium)}} \\

 Alg $\backslash$ Task & Box Close & Coffee Pull & Coffee Push & Hammer & Peg Insert Side & Push Wall & Soccer \\
\midrule

 3D Diffusion Policy & \dd{18}{8} & \dd{52}{23} & \dd{55}{0} & \ddbf{77}{6} & \dd{58}{6} & \dd{60}{8} & \dd{8}{5} \\
 3D Flow Matching* & \dd{18}{8} & \dd{67}{2} & \dd{58}{15} & \dd{75}{22} & \dd{68}{5} & \dd{15}{15} & \ddbf{10}{4} \\
\textbf{3D \ours} & \ddbf{45}{12} & \ddbf{97}{2} & \ddbf{82}{6} & \dd{42}{24} & \ddbf{88}{8} & \ddbf{93}{2} & \dd{7}{6} \\

\bottomrule
\end{tabular}}

\resizebox{1.0\textwidth}{!}{%
\begin{tabular}{l|ccccccc}
\toprule
& \multicolumn{2}{c|}{\textbf{Meta-World (Medium)}} & \multicolumn{5}{c}{\textbf{Meta-World (Hard)}} \\

 Alg $\backslash$ Task & Sweep & Sweep Into & Assembly & Hand Insert & Pick Out of Hole & Pick Place & Push \\
\midrule

 3D Diffusion Policy & \dd{70}{4} & \dd{3}{2} & \dd{77}{16} & \dd{7}{9} & \ddbf{20}{11} & \dd{42}{5} & \dd{55}{14} \\
 3D Flow Matching* & \dd{63}{21} & \dd{0}{0} & \dd{88}{10} & \dd{0}{0} & \dd{38}{2} & \dd{53}{17} & \dd{62}{2} \\
\textbf{3D \ours} & \ddbf{92}{2} & \ddbf{7}{2} & \ddbf{100}{0} & \ddbf{12}{9} & \dd{13}{5} & \ddbf{68}{5} & \ddbf{88}{9} \\

\bottomrule
\end{tabular}}

\resizebox{0.8\textwidth}{!}{%
\begin{tabular}{l|cccc|c}
\toprule
& \multicolumn{4}{c|}{\textbf{Meta-World (Very Hard)}} & 
 \multicolumn{1}{c}{\multirow{2}{*}{\textbf{Average}}} \\

 Alg $\backslash$ Task & Shelf Place & Disassemble & Stick Pull & Pick Place Wall &  \\
\midrule

 3D Diffusion Policy & \dd{25}{8} & \dd{55}{19} & \dd{28}{14} & \dd{55}{27} & \multicolumn{1}{c}{\dd{59.4}{3.5}}\\
 3D Flow Matching* & \dd{18}{10} & \ddbf{67}{5} & \dd{43}{25} & \dd{40}{11} & \multicolumn{1}{c}{\dd{57.9}{0.5}}\\
\textbf{3D \ours} & \ddbf{28}{5} & \dd{63}{8} & \ddbf{83}{5} & \ddbf{98}{2} & \multicolumn{1}{c}{\ddbf{78.1}{2.0}}\\

\bottomrule
\end{tabular}}
\end{flushleft}
\vspace{-0.2in}
\end{table*}